\documentclass{article}

\usepackage{microtype}
\usepackage{graphicx}
\usepackage{booktabs} %

\usepackage{hyperref}

\usepackage{amsmath,amsfonts,bm}

\def\eqref#1{equation~\ref{#1}}

\def\1{\bm{1}}

\DeclareMathAlphabet{\mathsfit}{\encodingdefault}{\sfdefault}{m}{sl}
\SetMathAlphabet{\mathsfit}{bold}{\encodingdefault}{\sfdefault}{bx}{n}

\usepackage{hyperref}

\usepackage{url}
\usepackage{multirow}
\usepackage{rotating}
\usepackage{float}

\usepackage{booktabs}
\usepackage{multirow}
\usepackage{tabularx}
\usepackage{booktabs}
\usepackage{tabularray}
\usepackage{graphics}
\usepackage{amssymb}
\usepackage{pifont}
\usepackage{lipsum}
\usepackage{subcaption}
\usepackage{comment}
\usepackage[table]{xcolor}
\usepackage[title]{appendix}
\usepackage[accsupp]{axessibility}
\usepackage{enumitem}
\usepackage{comment}
\usepackage{titletoc}

\usepackage[arxiv]{icml2025}

\usepackage{amsmath}
\usepackage{amssymb}
\usepackage{mathtools}
\usepackage{amsthm}

\usepackage[capitalize,noabbrev]{cleveref}

\usepackage{ifthen}
\newboolean{arxiv}
\setboolean{arxiv}{true}
\newboolean{icmlfinal}
\setboolean{icmlfinal}{false}

\theoremstyle{plain}

\theoremstyle{definition}

\theoremstyle{remark}

\usepackage[textsize=tiny]{todonotes}

\newcommand{\revision}[1]{{\color{black}#1}}
\newcommand{\model}{UniVLG}
\newcommand{\Det}{\texttt{Det}}
\newcommand{\GT}{\texttt{GT}}

\definecolor{three_d_color}{HTML}{dab543}
\definecolor{two_d_color}{HTML}{087ca6}

\icmltitlerunning{Unifying 2D and 3D Vision-Language Understanding}

\begin{document}

\twocolumn[
\icmltitle{Unifying 2D and 3D Vision-Language Understanding}
\icmlsetsymbol{equal}{*}

\begin{icmlauthorlist}
\icmlauthor{Ayush Jain}{equal,cmu,meta}
\icmlauthor{Alexander Swerdlow}{equal,cmu}
\icmlauthor{Yuzhou Wang}{cmu}
\icmlauthor{Sergio Arnaud}{meta}
\icmlauthor{Ada Martin}{meta}
\icmlauthor{Alexander Sax}{meta}
\icmlauthor{Franziska Meier}{meta}
\icmlauthor{Katerina Fragkiadaki}{cmu}
\end{icmlauthorlist}

\icmlaffiliation{cmu}{Carnegie Mellon University}
\icmlaffiliation{meta}{Meta Inc.}

\icmlcorrespondingauthor{Ayush Jain}{ayushj2@andrew.cmu.edu}
\icmlcorrespondingauthor{Alexander Swerdlow}{aswerdlo@stanford.edu}

\icmlkeywords{Machine Learning, ICML}

\vskip 0.3in
]

\ifthenelse{\boolean{arxiv} \or \boolean{icmlfinal}}{
\printAffiliationsAndNotice{\icmlEqualContribution}
}{}

\newcommand{\fix}{\marginpar{FIX}}
\newcommand{\new}{\marginpar{NEW}}
\newcommand{\commentcustom}[1]{}

\begin{abstract}
Progress in 3D vision-language learning has been hindered by the scarcity of large-scale 3D datasets. We introduce \model{}, a unified architecture for 2D and 3D vision-language understanding that bridges the gap between existing 2D-centric models and the rich 3D sensory data available in embodied systems. Our approach initializes most model weights from pre-trained 2D models and trains on both 2D and 3D vision-language data. We propose a novel language-conditioned  mask decoder shared across 2D and 3D modalities to ground objects effectively in both RGB and RGB-D images, outperforming box-based approaches. To further reduce the domain gap between 2D and 3D, we incorporate 2D-to-3D lifting strategies, enabling \model{} to utilize 2D data to enhance 3D performance. With these innovations, our model achieves state-of-the-art performance across multiple 3D vision-language grounding tasks, demonstrating the potential of transferring advances from 2D vision-language learning to the data-constrained 3D domain. Furthermore, co-training on both 2D and 3D data enhances performance across modalities without sacrificing 2D capabilities. By removing the reliance on 3D mesh reconstruction and ground-truth object proposals, \model{} sets a new standard for realistic, embodied-aligned evaluation. Code and additional visualizations are available at \href{https://univlg.github.io}{univlg.github.io}.
\end{abstract}

\begin{figure*}[h!]
\centering
    \includegraphics[width=\textwidth]{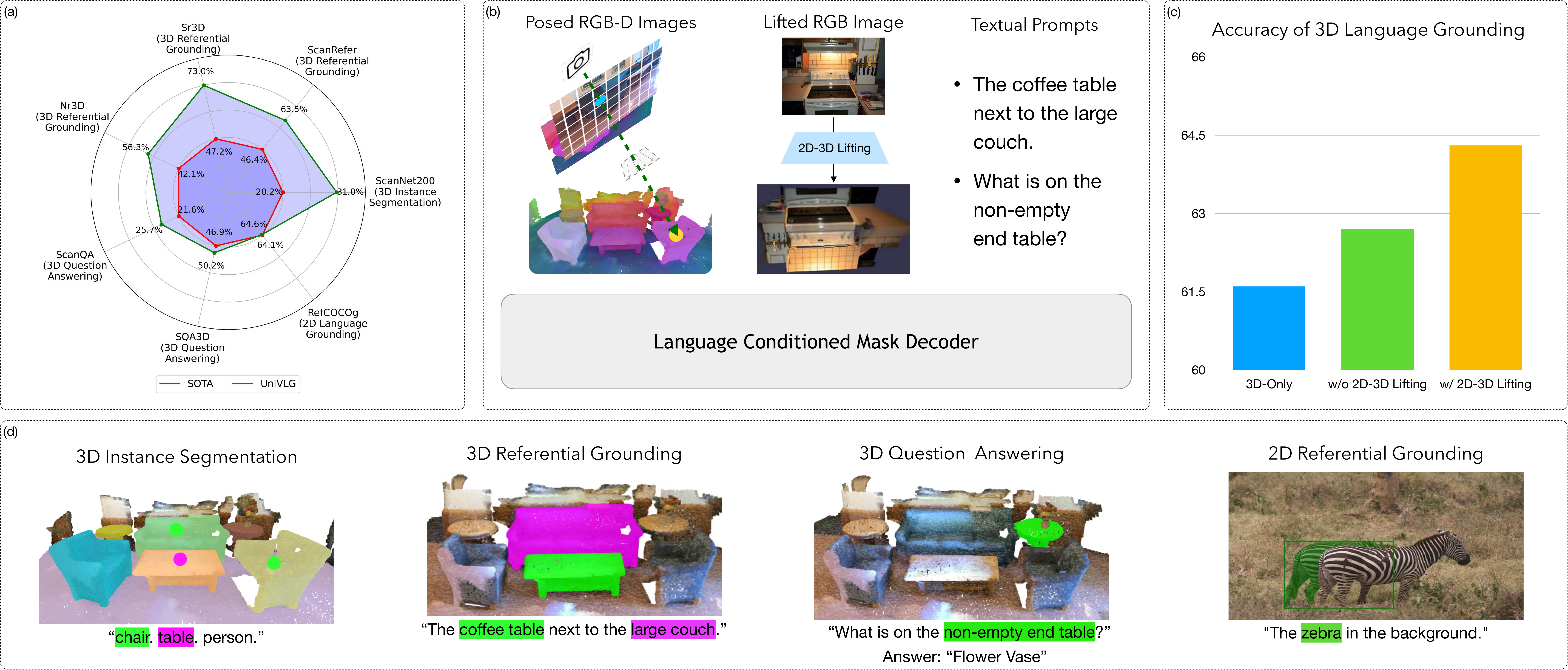}
    \caption{\textbf{(A)} \model{} achieves state-of-the-art performance performance across a range of referential grounding, question answering, and instance segmentation benchmarks. \textbf{(B)} \model{} is a unified model which accepts posed RGB-D sequences, or monocular 2D images which are then lifted to 3D pointmaps. \textbf{(C)} \model{} significantly benefits from joint 2D-3D training, further boosted when all parameters are shared between modalities by using 2D-3D lifting. \textbf{(D)} Example task inputs/outputs for \model{}.
    }
    \label{fig:teaser}
\end{figure*}

\vspace{-2em}
\section{Introduction}
Today’s real-world embodied systems rely on depth sensors and egocentric, calibrated camera setups for navigation and interaction with their surroundings \cite{saycan,mobilityvla}. However, despite having access to rich 3D information, these systems predominantly use 2D vision-language models to interpret their sensory video input, rather than leveraging 3D models that incorporate depth and egomotion. 
At first glance, this reliance on 2D models appears counterintuitive, as prior research has consistently shown that 3D models outperform their 2D counterparts when trained on comparable amounts of data \cite{pl,kundu2020virtualmultiviewfusion3d,move,imvoxelnet}. The key limitation, however, is dataset availability: while 2D datasets are vast and well-curated, 3D datasets remain scarce and expensive to annotate \cite{scannet,scannetpp}. As a result, there are currently no high-performing, pre-trained 3D encoders capable of processing  3D inputs at the same level as CLIP \cite{clip} does for 2D images. This data imbalance has led to a significant performance gap, ultimately slowing the widespread adoption of 3D models in embodied systems. 
Given these challenges, is scaling 3D training data the only viable path to bridging this gap, or are there alternative strategies for making 3D models more effective?

In this paper, we introduce \model{}, a unified 2D-3D vision-language model designed to improve 3D understanding by leveraging large-scale 2D data and pre-trained 2D models.  \model{} is trained on both 2D and 3D vision-language tasks, including referential grounding, object detection, and question answering in images and 3D scenes.
Unlike models that operate directly on 3D point clouds, \model{} processes RGB and RGB-D images---natural sensory inputs for embodied agents---and supports both single-view RGB images or multi-view posed RGB-D images. \model{} processes each image with strong pre-trained 2D backbones, which also constitute the majority of its parameters, and fully leverages their representational power.  It  discriminates between 2D and 3D purely through the positional encodings of 2D image patch features, which represent the 2D pixel grid locations in images and the 3D (X,Y,Z) coordinates in scenes, similar to \cite{odin}. When training on 2D RGB images, we consider both 2D and 3D processing pathways within \model{}, by using predicted 3D pointmaps 
~\cite{moge}, which further narrows the domain gap between 2D and 3D input. We further introduce a novel language-conditioned  mask decoder, shared across both 2D and 3D input, which predicts segmentation masks by conditioning on both visual features and language instructions to  ground objects mentioned in the language input. Segmentation masks serve as a unifying output representation because they involve per-patch predictions, where each patch corresponds to either a 2D pixel or a 3D point. In our experiments, we show that besides unifying the output space, decoding to 3D masks results in significantly more precise predictions, challenging the established paradigm which decode bounding boxes or rely on object proposals. Our model is designed with the goal of sharing all weights across RGB image and RGB-D image sequence processing.

We test \model{} on established 2D and 3D vision language benchmarks \cite{referit3d, scanrefer}. 
We find that when trained exclusively on 3D data, \model{} achieves state-of-the-art performance across all established benchmarks, outperforming prior methods in comparable settings by more than 15\%. Furthermore, co-training \model{} with 2D data enhances its 3D performance even further, both on in-domain and out-of-domain benchmarks. Notably, this improvement does not come at the expense of 2D tasks—\model{} retains strong performance on 2D referential grounding datasets~\cite{refcoco} compared to its version which is only trained on 2D referential grounding data. 
\model{} directly uses sensor point clouds without any mesh pre-processing of the RGB-D input and without relying on ground-truth bounding box proposals, typically used in existing works and benchmarks \cite{referit3d}. 
By benchmarking in these more realistic settings, we hope to encourage future research that aligns more closely with the goals of embodied vision and promotes the progress of 3D vision in practical, real-world scenarios.

In summary, our contributions are:
\begin{itemize}[itemsep=1pt, left=0.3cm, topsep=0pt]
   \item \textbf{Unified 2D-3D Visual Grounding}: We propose a model that can consume and benefit from both 2D and 3D vision-language data. 

    \item \textbf{State-of-the-Art Performance:}  \model{} achieves state-of-the-art performance on in-domain 3D referential grounding benchmarks, including ReferIt3D (SR3D, NR3D) and ScanRefer, outperforming prior methods by a significant margin, while also excelling in out-of-domain 3D referential grounding datasets.
      \item \textbf{Language-Conditioned 3D Mask Decoder:} We propose a novel language-conditioned  mask decoder head for 3D referential grounding and show its superior performance over bounding box decoders. 
      \item \textbf{Realistic Evaluation Settings:} We benchmark prior methods and \model{} in  realistic, embodied-aligned settings by using sensor-generated instead of mesh-reconstructed point clouds and without relying on ground-truth object proposals. 
\end{itemize}

We make our code publicly available
at \href{https://univlg.github.io}{univlg.github.io}.

\vspace{-6pt}
\section{Related Work}

\paragraph{Use of 2D data for 3D Visual Language Understanding Tasks} Most 3D Visual Language models directly operate over the provided 3D point clouds without using any 2D pre-trained features. SAT-2D \cite{Yang2021SAT2S} is one of the first 3D visual grounding model which used 2D visual features during training for aligning 2D and 3D visual features and show significant boost over its versions that do not use 2D features. Recent methods in 3D Question Answering like 3D-LLM \cite{3dllm} and NaviLLM \cite{navillm} use multi-view 2D features and pass them to LLMs for decoding answers. However, so far they haven't been able to successfully address 3D visual grounding tasks. PQ3D \cite{pq3d} uses a combination of several visual backbones, including a 2D based feature backbone from OpenScene \cite{openscene}. Recent work of EFM3D \cite{straub2024efm3dbenchmarkmeasuringprogress} uses 3D feature volumes obtained from lifting 2D image features but only evaluates on the task of 3D object detection and surface reconstruction.

Another related line of research focuses on enhancing 2D vision-language models (VLMs) with 3D reasoning. Spatial-RGPT \cite{spatialrgpt} and Spatial-VLM \cite{spatialvlm} use depth estimation to enrich 2D models with spatial understanding. While these methods focus on improving 2D perception, our approach leverages 2D-to-3D lifting to enhance multi-view 3D reasoning, bridging the gap between 2D and 3D for vision-language grounding.

The closest work to ours is ODIN \cite{odin}, which also differentiates between 2D and 3D through positional encodings instead of using separate image and point cloud encoders. However, they only consider the task of object segmentation.   
\model{} is inspired by ODIN and innovates over it in the following ways: a) It extends its applicability to referential grounding and question-answering tasks. b) 
It improves the mask decoder to better incorporate language information.  c) It  shares all parameters between 2D and 3D pathways, instead of a subset of them by lifting 2D images to 3D pointmaps. With these advancements, \model{} dramatically outperforms ODIN, and its extension  LLaVA-3D \cite{l3d} on 3D language grounding, demonstrating the importance of its design choices.

\vspace{-1em}
\paragraph{3D Visual Grounding Models}
3D Visual Grounding Models can be broadly divided into two categories: Two-stage methods and single-stage end-to-end methods. Two stage methods first generate 3D object proposals and then select one proposal out of them. This is the dominant paradigm: InstanceRefer \cite{Yuan2021InstanceReferCH}, SAT-2D \cite{Yang2021SAT2S}, ViL3DRel \cite{vil3d} and recently scaled-up to models of 3D-VisTA \cite{3dvista} and PQ3D \cite{pq3d} which train their model on multiple 3D datasets and tasks. Specifically, 3D-VisTA first pre-trains their model on masked language/object modeling and scene-text matching, and then fine-tunes to downstream  several language understanding  tasks of interest. PQ3D \cite{pq3d} %
proposes promptable object queries for  3D scene understanding. While it decodes masks for instance segmentation tasks directly, it follows a 2D stage approach for free-form language grounding  and selects a mask from a set of  object mask proposals. However, two-stage methods are limited by the failures of the object proposal networks. To overcome this limitation, single-stage methods like 3D-SPS \cite{3dsps} and BUTD-DETR \cite{butd} directly regress 3D bounding boxes. They achieve strong results, especially on benchmarks like ScanRefer, which do not provide ground-truth proposals. However, they have only been trained on individual tasks and datasets and have not been scaled up yet. In this work, we propose a single-stage end-to-end model that is jointly trained on multiple 3D language understanding tasks, and achieve state-of-the-art results on several benchmarks. 

For additional related work, see Appendix (Section-\ref{appendix:related_work}).

\section{Method}

\begin{figure*}[ht]
     \centering
     \includegraphics[width=\textwidth]{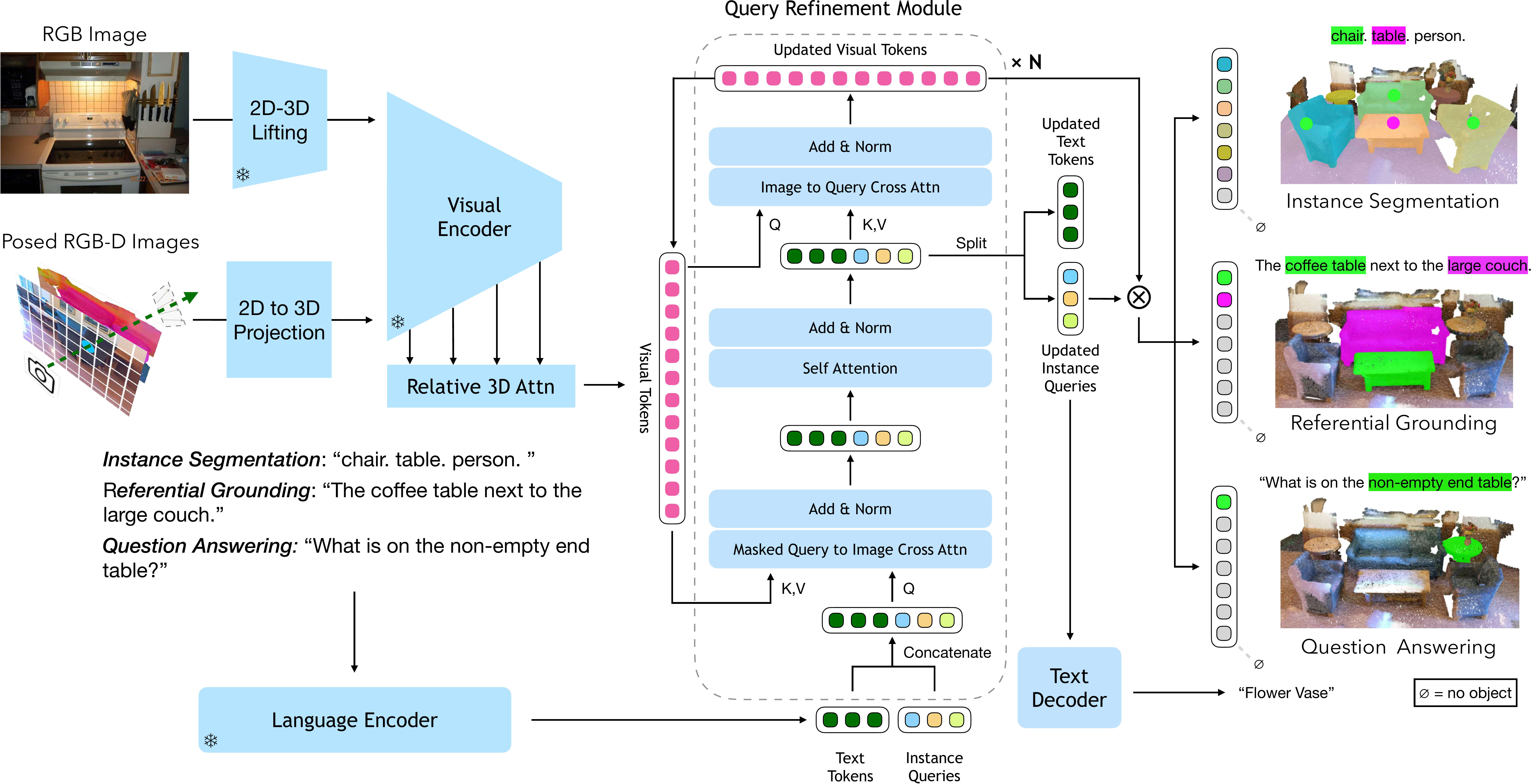}
    \caption{ \footnotesize{\textbf{\model{} Architecture}:
    A vision language transformer that accepts a language utterance and either (1) a sequence of posed RGB-D images or (2) a monocular RGB image, lifted to 3D (2D to 3D Projection). \model{} fuses information across vision and language to predict 3D object segments or generate answers. It uses a ViT backbone followed by 3D relative attentions to produce a set of 3D feature tokens. The proposed decoder then iteratively updates a set of learnable queries as well as the 3D feature tokens though token - language - query attentions to decode object segments and match them to noun phrases in the input referential utterance. Masks are decoded through a dot-product between 3D feature tokens and learnable queries. A text decoder predicts answers for the input questions by conditioning on the set of updated object queries. }}
     \label{fig:method}
\vspace{-9pt}
\end{figure*}

We show the architecture of \model{} in Figure-\ref{fig:method}. The model takes as input a language query, $N$ RGB images of shape $N \times H \times W \times 3$, and an associated 3D pointmap of shape $N \times H \times W \times 3$. The output consists of segmentation masks for each object mentioned in the sentence, a corresponding text span that refers to each segmented object, and optionally, generated text that answers the question. In datasets such as ScanNet, we obtain the 3D pointmap by unprojecting the sensed depth images using the camera parameters and standard pinhole-camera equations. For RGB images from 2D datasets like RefCOCO~\cite{refcoco}, we use a neural 2D-to-3D lifting model~\cite{moge}, which takes a (monocular) RGB image as input and predicts a 3D pointmap. Note that the 3D pointmap does not need to be metric—in fact, our 3D pointmaps for 2D datasets are represented in relative space.

\textbf{Visual Encoder:} We encode each RGB image independently using DiNO VIT encoder~\cite{oquab2024dinov2learningrobustvisual}, and add several 3D attention layers~\cite{odin} on top of features from multiple layers. Specifically, we apply 3D $k$-NN attention with relative positional embeddings to fuse information across the input RGB views. This attention mechanism uses feature maps from the ViT encoder, with 3D pointmaps serving as the positional embeddings. Since our attention mechanism is relative, our model does not require a metric pointmap.

\textbf{Language Encoder:} We embed the natural language query using JinaCLIP~\cite{koukounas2024jina}, generating tokens of shape $M \times F$ where $M$ is the number of tokens and $F$ is the feature dimension.

\textbf{Language Conditioned Mask Decoder:} 
The mask decoder head takes as input the encoded visual features, their corresponding (relative) 3D coordinates, and the encoded language utterance; it outputs 3D segmentation masks of the mentioned objects and a text span over the encoded language utterance. Our mask decoder head draws inspiration from Mask2Former \cite{m2f} and makes important architectural changes to make it suitable for 3D referential grounding.

We initialize a set of $M$ learnable object queries, each responsible for decoding an object instance. We concatenate these object queries with the language tokens along the sequence dimension. We alternate between cross-attention between these and the visual tokens and self-attention among these concatenated queries and text tokens. Instead of using a vanilla cross-attention layer, we follow Mask2Former and use a masked variant where each query only attends to the points falling within the corresponding instance mask predicted by the previous layer. For this operation, we add 3D positional embeddings on the visual features. 
Next, the visual tokens from the backbone are updated by cross-attending to the updated object and text tokens. 
Specifically, let $Q^{(0)} \in \mathbb{R}^{M \times D}$ be the initial object queries, $T \in \mathbb{R}^{L \times D}$ be the text tokens, and $V^{(0)} \in \mathbb{R}^{N \times D}$ be the 3D visual tokens. The query refinement process can be described as:
\vspace{-20pt}
\begin{align*}
    X^{(0)} &= [Q^{(0)}; T]; \\
    X^{(i+1)} &= \text{Norm}(\text{MaskedCrossAttention}(X^{(i)}, V^{(i)}) + X^{(i)}) \\
    X^{(i+1)} &= \text{Norm}(\text{SelfAttention}(X^{(i+1)}) + X^{(i+1)}) \\
    V^{(i+1)} &= \text{Norm}(\text{CrossAttention}(V^{(i)}, X^{(i+1)}) + V^{(i)})),
\end{align*}
where $[;]$ denotes concatenation along the sequence dimension, and $i$ is the layer index. The refined queries after each decoder layer $Q^{(i+1)} = X^{(i+1)}_{1:M}$ are then used for mask prediction with the updated visual features and for language grounding.

We find that \textit{updating visual features} via attention to queries and text tokens \textit{is crucial for 3D-referential grounding}. Open-vocabulary mask decoders, such as those in ODIN~\cite{odin} and X-Decoder~\cite{xdecoder}, which extend Mask2Former's decoder to accept language tokens, do not update visual features during query refinement as in our method. Although their approach is sufficient for 3D instance segmentation, our experiments show that this choice significantly hinders performance in decoding object masks for 3D referential grounding (Table~\ref{table:ablations}). Object2Scene~\cite{object2scene}, which decodes 3D bounding boxes for referential grounding, finds that only updating queries is sufficient. However, our ablation studies show that while this holds true for bounding box decoding, updating visual features during query refinement is crucial for accurately decoding masks (Table~\ref{tab:feat_attn}).

After attending to text and visual features, the refined object queries decode object segments through a token-wise dot-product with the updated visual features to produce mask logits which are then thresholded to obtain segmentation masks:
\vspace{-3pt}
\begin{equation}
    M_i = \sigma(\text{sigmoid}(Q^{(f)}_i \cdot V^T)),
\end{equation}
\vspace{-1pt}
where $M_i$ is the mask for the $i$-th object query, $\sigma$ is a threshold function, and $\cdot$ denotes dot product.

\textbf{Text Decoder:} 
Beyond decoding segments, the refined object queries are used as input to the decoder of a pre-trained T5~\cite{raffel2020exploring} decoder to generate answers to questions, following PQ3D~\cite{pq3d}. This is useful for question-answering tasks where the output is a text sentence. %

\vspace{-6pt}
\subsection{Supervision Objective}
\textbf{Mask Loss}: We match queries to ground-truth instances using Hungarian Matching \cite{Carion2020EndtoEndOD}. We supervise the matched queries's predicted masks with both a Binary Cross Entropy (BCE) and Dice loss following Mask2Former~\cite{m2f}.

\vspace{-2pt}
\textbf{Text Span Loss:} Similar to prior works \cite{glip, Kamath2021MDETRM, butd}, we match the predicted 3D object segmentations to the relevant noun phrases in the input utterance through a dot-product between the object queries and the language tokens, generating the distribution $G_i$ over the input text sentence for the $i$th query: 
\vspace{-1pt}
\begin{equation}
    G_i = \text{sigmoid}(f_\phi (Q^{(f)}_i) \cdot f_\theta (T^T))
\end{equation}
\vspace{-1pt}
where $f_\phi$ and $f_\theta$ are MLPs, $G_i$ is the grounding distribution for the $i$-th object query over the input text tokens. We supervise these grounding distributions with a BCE loss, with unmatched queries supervised to have a low probability.

\vspace{-2pt}
\textbf{Box Loss:} We observe a failure mode in our model where, when trained with the aforementioned objectives, some masks include a small number of distant, unrelated points, or multiple instances of the same object category are predicted by a single object query (see Figure~\ref{fig:failure} in Appendix). To address this, we introduce a novel box loss. This loss computes an enclosing 3D bounding box for each predicted mask and supervises it using standard box prediction losses—L1 and Generalized Intersection-over-Union (GIoU)~\cite{rezatofighi2019generalized}—against the ground-truth bounding boxes. We incorporate this box loss as an additional cost in both Hungarian matching and the final loss. This encourages the model to produce more accurate and compact masks, leading to improved downstream performance.

\vspace{-2pt}
\textbf{Text Generation Loss: } For question answering tasks, our model decodes a text utterance as an output. We supervise the generated text with the ground-truth text answer using standard cross-entropy loss. In summary, our complete loss function is as follows:

{\vspace{-1.75em}
\begin{equation}
    \mathcal{L}_{\text{total}} = \lambda_{\text{mask}}\mathcal{L}_{\text{mask}} + \lambda_{\text{text}}\mathcal{L}_{\text{text}} + \lambda_{\text{box}}\mathcal{L}_{\text{box}} + \lambda_{\text{gen}}\mathcal{L}_{\text{gen}}
\end{equation}
\vspace{-1.75em}}

\revision{where $\mathcal{L}_{\text{mask}}$ is the mask loss comprised of binary cross entropy and dice losses, $\mathcal{L}_{\text{text}}$ is the loss for matching the object queries to the mentioned objects in the language sentence, $\mathcal{L}_{\text{box}}$ are the additional bounding box losses described earlier, and $\mathcal{L}_{\text{gen}}$ is the cross-entropy loss over the auto-regressively generated answer (in case of question-answering datasets).} 

\model{} shares all learnable parameters between 2D and 3D by leveraging 2D-to-3D lifting strategies to generate pointmaps for 2D datasets, which are seamlessly integrated into 3D attention layers. Additionally, the mask decoding head unifies the output space between 2D and 3D as per-pixel segmentation masks, enabling the sharing of both loss functions and decoder parameters across modalities.

\noindent \textbf{Implementation details:}
\model{} consists of 108M trainable parameters along with a frozen 220M parameter text-encoder~\cite{koukounas2024jina} and a 304M parameter image-encoder~\cite{oquab2024dinov2learningrobustvisual}. For ablations in \cref{table:ablations} and \ref{table:box_mask}, we use a 88M parameter Swin~\cite{swin} image-encoder. We train in data-parallel across 32 A100 80G GPUs with an effective batch size of 64. We use ScanEnts3D \cite{scanents} version of ScanRefer~\cite{scanrefer} and Referit3D~\cite{referit3d} which provides object annotations for all noun words in the language sentence. During training, we process either a sequence of $N$ posed RGB-D images, or a single RGB image. During training, the model processes either a sequence of $N$ posed RGB-D images or a single RGB image. For 2D images, we apply a 2D-to-3D lifting strategy with a 50\% probability. When lifted, the images pass through all 2D-3D layers; otherwise, they remain in 2D space, skipping the 3D attention layers. At test time, we retain 2D images in their original space to prevent noise from predicted 3D pointmaps from impacting 2D performance. For 3D scenes, we compute CLIP embeddings for all images and captions and use this to select 5 relevant frames, with an additional 10 frames coming from Furthest-Point-Sampling (FPS) in the CLIP embedding space, for a total of 15 frames. At test time, we feed all images in a scene to our model. For validation results, we perform span prediction to identify the primary subject from a given utterance. We found that prompting an LLM~\cite{Dubey2024TheL3} with examples specific to a given dataset to result in better performance compared to traditional NLP libraries. We use Jina-CLIP~\cite{koukounas2024jina} as the text-encoder, as it supports arbitrary input-length. We jointly train our model on all datasets, with text generation loss only active in question answering datasets. Our method provides for fast inference, with a \textit{90-frame} scene taking $\sim$1050ms and $\sim$15GB of VRAM on an A100 GPU.

\vspace{-10pt}
\section{Experiments}
\vspace{-3pt}

\begin{table*}[tb]
    \centering
    \vspace{-0.5em}
    \caption{\textbf{Results on 3D language grounding in 3D mesh and sensor point clouds (PC).} We evaluate top-1 accuracy on the official validation set with assuming ground-truth (\GT) or without assuming ground-truth proposals (\Det).}
    \vspace{2pt}
    \resizebox{.98\textwidth}{!}
    {
    \begin{tabular}{ll*{11}{c}}
        \toprule
        & & \multicolumn{4}{c}{\textbf{SR3D}} & \multicolumn{4}{c}{\textbf{NR3D}} & \multicolumn{3}{c}{\textbf{ScanRefer}}\\
        & \textbf{Method} & \begin{tabular}{@{}c@{}}Acc \\ @25\\ (\Det)\end{tabular} & \begin{tabular}{@{}c@{}}Acc \\ @50\\ (\Det)\end{tabular} & \begin{tabular}{@{}c@{}}Acc \\ @75\\ (\Det)\end{tabular} & \begin{tabular}{@{}c@{}}Acc \\ (\GT)\end{tabular} & \begin{tabular}{@{}c@{}}Acc \\ @25\\ (\Det)\end{tabular} & \begin{tabular}{@{}c@{}}Acc \\ @50\\ (\Det)\end{tabular} & \begin{tabular}{@{}c@{}}Acc \\ @75\\ (\Det)\end{tabular} & \begin{tabular}{@{}c@{}}Acc \\ (\GT)\end{tabular} & \begin{tabular}{@{}c@{}}Acc \\ @25\\ (\Det)\end{tabular} & \begin{tabular}{@{}c@{}}Acc \\ @50\\ (\Det)\end{tabular} & \begin{tabular}{@{}c@{}}Acc \\ @75\\ (\Det)\end{tabular} \\
        \midrule
        \multirow{10}{0.8cm}{Mesh PC} & 
        ReferIt3DNet~\cite{referit3d} & 27.7 & - & - & 39.8 & 24.0 & - & - & - & 26.4 & 16.9 & - \\
        & ScanRefer~\cite{scanrefer}  & - & - & - & - & - & - & - & - & 35.5 & 22.4 & -\\
        & InstanceRefer~\cite{yuan2021instancerefer} & 31.5 & - & - & 48.0 & 29.9 & - & - & - & 40.2 & 32.9 & -\\
        & LanguageRefer~\cite{roh2022languagerefer}  & 39.5 & - & - & 56.0 & 28.6 & - & - & - & - & - & -\\
        & SAT-2D~\cite{yang2021sat} & 35.4 & - & - & 57.9 & 31.7 & - & - & - & 44.5 & 30.1& -\\
        & BUTD-DETR~\cite{butd} & 52.1 & - & - & 67.0 & 43.3 & - & - & 54.6 & 52.2 & 39.8 & - \\
        
        & 3D-VisTA~\cite{3dvista} & 56.5 & 51.5 & 42.8 & 76.4 & 47.7 & 42.2 & 35.5 & 65.1 & 51.0 & 46.2 & 36.7\\
        & LLaVA-3D~\cite{l3d} & - & - & - & - & - & - & - & - & 54.1 & 42.2 & - \\ 
        & PQ3D~\cite{pq3d} &\textbf{ 62.0} & \textbf{55.9} &\textbf{ 46.2} & \textbf{79.7} & \textbf{52.2} & \textbf{45.0 }& \textbf{37.6} &\textbf{ 66.7} & \textbf{56.7} & \textbf{51.8 }& \textbf{43.3}\\
        \midrule
        \multirow{5}{0.8cm}{Sensor PC}
        & ODIN~\cite{odin} & 38.1 & 29.3 & 23.1 & - & 31.6 & 20.8 & 15.8 & - & 43.1 & 33.4 & 26.2 \\
        & BUTD-DETR~\cite{butd} & 43.3 & 28.9 & 6.58 & - & 32.2 & 19.4 & 3.64 & - & 42.2 & 27.9 & 6.53 \\
        & 3D-VisTA~\cite{3dvista} & 47.2 & 43.2 & 36.1 & 61.4 & 42.1 & 37.4 & 32.0 & 54.2 & 46.4 & 42.5 & 36.3 \\
        & 
        \model-3D-only\;(Ours) & 71.6 & 63.8 & 49.4 & \textbf{81.7} & 54.7 & 44.9 & 35.7 & \textbf{65.2} & 60.7 & 53.2 & 42.6 \\
        &

        \model\;(Ours) & \textbf{73.0} & \textbf{64.8} & \textbf{51.8} & - & \textbf{58.3} & \textbf{49.8} & \textbf{39.1} & - &\textbf{ 63.5} & \textbf{56.4} & \textbf{46.0} \\
        
        \bottomrule 
    \end{tabular}}
    \label{table:grounding}
    \vspace{-1.0em}
\end{table*}

We evaluate our model on 3D and 2D referential grounding, 3D question answering and 3D instance segmentation benchmarks. We train our model on the 3D referential grounding datasets of SR3D, NR3D \cite{referit3d} and ScanRefer \cite{scanrefer} and 3D instance segmentation datasets of ScanNet200 \cite{scannet200} and Matterport3D \cite{matterport}. In addition to the 3D datasets, we also train our model on 2D referential grounding datasets with RefCOCO, RefCOCO+ and RefCOCOg \cite{refcoco}, and 2D image segmentation dataset with COCO \cite{coco}. We present results for two model versions: one trained solely on 3D data (\model{}-3D-only) and the other trained jointly on both 2D and 3D datasets (\model{}).

\vspace{-7pt}
\subsection{Evaluation on 3D Referential Grounding}
\vspace{-3pt}
Following BUTD-DETR \cite{butd}, we test on two evaluation setups: 
1. $\Det$, where our model and baselines do not have access to ground-truth 3D boxes of objects in the scene, and 2. $\GT$,  where our model and baselines use ground-truth 3D object proposals provided in the benchmarks. 

Additionally, these benchmarks sample point clouds from reconstructed and post-processed meshes instead of directly using the raw point clouds obtained from sensor inputs. As observed in prior works~\cite{odin,kundu2020virtualmultiviewfusion3d}, mesh-sampled point clouds often exhibit fine-grained misalignment with sensor-generated point clouds, which can unfairly disadvantage sensor-based approaches on these benchmarks. To address this, we benchmark our method and prior methods in a more embodied-aligned setting using sensor point clouds directly.
We thus evaluate all methods on benchmark-provided point clouds sampled from the post-processed mesh (\textit{Mesh}), and separately retrain and evaluate a subset of methods on sensor point clouds (\textit{Sensor}) obtained by unprojecting posed RGB-D images.

\noindent \textbf{Evaluation Metrics:} We use the standard top-1 accuracy metric. For the \Det{} setup, a predicted bounding box is considered correct if its intersection over union (IoU) with the ground truth box is higher than a predetermined threshold (we use the standard $0.25$, $0.5$ and $0.75$). As \model{} predicts masks (instead of axis-aligned bounding boxes), we obtain a bounding box by taking the extents of the mask. For the \GT{} setup, we pool visual features inside the given ground-truth masks, and the object queries predict a segmentation mask over the ``pooled" feature tokens, one token per object. The prediction is correct if the model selects the feature token corresponding to the ground-truth object.

\noindent \textbf{Baselines:} We compare our model against the state-of-the-art two-stage methods of 3D-VisTA \cite{3dvista}, PQ3D \cite{pq3d} and concurrent work of LLaVA-3D \cite{l3d}; and the SOTA single-stage method of BUTD-DETR \cite{butd}. \model{} uses significantly less \textit{3D} training data than prior SOTA 3D referential grounding models. For example, 3D-VisTA \cite{3dvista} trains on the previously mentioned 3D datasets that we use but also includes 3RScan (1500 scenes) \cite{3rscan}, Objaverse (700k objects) \cite{objaverse}, and additional text sentences on ScanNet generated using GPT-3 (see Table-3 of 3D-VisTA). Similarly, PQ3D adds the Multi3DRefer \cite{multi3drefer} and Scan2Cap datasets \cite{scan2cap}, but also utilizes a point encoder that was trained on all 3D-VisTA datasets. All two-stage baselines assume access to ground-truth proposals at test-time in the SR3D and NR3D benchmarks; hence we re-evaluate them with predicted boxes coming from a SoTA object detector, Mask3D \cite{mask3d}. We also re-train 3D-VisTA and BUTD-DETR with sensor point clouds. Despite our best efforts, we could not manage to re-train PQ3D with sensor point clouds due to their use of multiple backbones, and multi-stage training strategies. We also compare our model with ODIN \cite{odin} trained for 3D language grounding using their architecture but our grounding losses. The 3D referential grounding results are presented in \cref{table:grounding}, from which we find:

\noindent \textbf{\model{} outperforms prior methods, regardless of data selection, on all setups which do not assume GT boxes.} Even without our joint 2D training strategy—and with less 3D data than prior methods—\model{}-3D-only significantly outperforms all prior methods. It dramatically outperforms alternative single stage models, such as BUTD-DETR, on the stricter IoU threshold of 0.75, thanks to predicting masks instead of bounding boxes—as we demonstrate later in \cref{tab:iou}. In the \GT{} setup as well, \model{} significantly outperforms 3D-VisTA and closely matches the performance of the recent work of PQ3D in the setup where PQ3D uses mesh point clouds, while \model{} operates over sensor point clouds.

\noindent \textbf{Co-training \model{} with 2D and 3D data enhances 3D performance} across all 3D vision-language grounding benchmarks, demonstrating that leveraging 2D data during training provides additional benefits beyond initializing the model with pre-trained 2D weights (row-5, sensor PC, ~\cref{tab:established_3d}).

\noindent \textbf{Performance of all prior SOTA models drop with sensor point cloud as input and without assuming GT boxes}: Both single-stage methods like BUTD-DETR and two-stage methods like 3D-VisTA have a performance drop of 5-15\% when using sensor RGB-D point clouds as input instead of mesh point-clouds. The sensor point cloud and mesh point clouds have fine-grained misalignment, resulting in this drop. Shifting from ground-truth box proposals to a more realistic setup of using predicted box proposals from a SOTA detector results in a drop of 15-20\% in accuracy. Nonetheless, even when \model{} uses sensor pointclouds (which as we showed above result in a 5-15\% accuracy drop on these benchmarks), it still outperforms the baselines that use \textit{mesh} point clouds as input.

Qualitative results of \model{} are in \cref{fig:vis} (Appendix).

\begin{table}[tb]
\centering
\caption{\textbf{Out-of-Domain 3D Referential Grounding} Acc@25 in \Det. From left-to-right, ScanNet++, HM3D, ARKitScenes, ScanNet (GT), ScanNet (SAMPro3D). See \cref{appendix:ood_data} for details.}
\label{table:ood_det}
\begin{tabular}{@{}p{2.7cm}@{\hskip 0.25em}c@{\hskip 0.25em}c@{\hskip 0.25em}c@{\hskip 0.25em}c@{\hskip 0.25em}c@{}}
\toprule
\textbf{Model} & \textbf{SN++} & \textbf{HM3D} & \textbf{ARKit} & \textbf{SN-GT} & \textbf{SN-SAM} \\
\midrule
\model{}-3D-only & 42.4 & 49.7 & 64.6 & 77.0 & 50.9 \\
\model{}  & \textbf{42.8} & \textbf{51.9} & \textbf{65.1} & \textbf{79.9} & \textbf{52.5} \\
\bottomrule
\end{tabular}
\end{table}

\vspace{-6pt}
\subsection{Evaluation on Out-of-Domain 3D Referential Grounding}
\vspace{-3pt}
We evaluate our model and baselines on L3DD \cite{locate3d}, an out-of-domain 3D language grounding dataset that spans ScanNet \cite{scannet}, ScanNet++ \cite{scannetpp}, ARKitScenes \cite{arkitscenes}, and HM3D \cite{hm3d}. L3DD allows us to assess the robustness of our model on new scenes, camera capture systems, and language instructions. We show the results of our model, both a 3D-only variant and our full model w/2D data + lifting in \cref{table:ood_det}. We find that our model outperforms prior methods on these out-of-domain datasets, achieving strong performance across the board.

\vspace{-6pt}
\subsection{Evaluation on 3D Question Answering}

\begin{table}[tb]
    \vspace{-1.1em}
    \centering
    \caption{\textbf{Results on 3D Visual Question Answering} on official validation sets. We evaluate top-1 exact match accuracy (EM@1).}
    \vspace{2pt}
    \label{table:vqa}
    \resizebox{.49\textwidth}{!}
    {
    \begin{tabular}{llcc}
        \toprule
        & \textbf{Method} & \textbf{ScanQA} & \textbf{SQA3D} \\
        \midrule
        \multirow{4}{*}{Mesh PC}
        & 3D-LLM (BLIP2-flant5)~\cite{3dllm} & 20.5 & -- \\
        & PQ3D~\cite{pq3d} & 21.0 & 47.0\\
        & 3D-VisTA~\cite{3dvista} & 22.1 & \textbf{47.5}\\
        & NaviLLM~\cite{navillm} & \textbf{23.9} & -- \\
        \midrule
        \multirow{2}{*}{Sensor PC}
        & 3D-VisTA~\cite{3dvista}  & 21.6 & 46.9\\
        & \model{} (Ours) & \textbf{25.7} & \textbf{50.2} \\
        \bottomrule
    \end{tabular}}
    \vspace{-1.5em}
\end{table}

\vspace{-3pt}
We test \model{} on 
ScanQA \cite{scanqa} and SQA3D \cite{sqa3d} question answering benchmarks. ScanQA \cite{scanqa} focuses on spatial relations. Alongside question-answer pairs, the dataset  includes annotations for the objects referenced in the question, and we supervise our model to predict these in addition to generating the answer. SQA3D \cite{sqa3d} provides pairs of  situation descriptions and questions regarding embodied scene understanding, navigation, common sense  and multi-hop reasoning, such as,  \textit{``looking for some food in the fridge"},\textit{ ``which direction should i go?"} and  the task is to generate the correct answer (\textit{``right"}).

\noindent \textbf{Evaluation Metrics:}
We use the established Exact Match (EM@1) metric, which measures if the generated answer  matches either of the two answer candidates provided by ScanQA, or the single ground-truth answer provided by SQA3D. We also report results with additional metrics in Appendix (\cref{table:vqa_detail}).

\noindent \textbf{Baselines:}
We compare against the LLM based methods of 3D-LLM \cite{3dllm} and NaviLLM \cite{navillm} which use BLIP2-flanT5 \cite{blip2} and Vicuna-7B \cite{vicuna} as their answer generation heads. We also compare with 3D-VisTA \cite{3dvista} and PQ3D \cite{pq3d} which use small decoder heads like T5-small \cite{raffel2020exploring}, similar to our approach.
We show results in \cref{table:vqa} on the validation sets of these benchmarks. \model{} outperforms all prior baselines on both benchmarks. We found that using sensor point clouds vs mesh point clouds does not result in a significant difference in performance in these benchmarks, likely because the models are evaluated on text generation instead of localization of objects as in 3D referential grounding and segmentation benchmarks. Additionally, we show results on 3D instance segmentation in \cref{app:3d_instance_seg}.

\vspace{-6pt}
\subsection{Evaluation on 2D Referential Grounding}
\begin{table}[tb]
    \centering
    \vspace{-1.0em}
    \caption{\textbf{Results on val sets of 2D Ref. grounding datasets}}
    \vspace{2pt}
    \resizebox{.49\textwidth}{!}
    {
    \begin{tabular}{ll*{2}{c}}
         \toprule
        & RefCOCO & RefCOCO+ & RefCOCOg \\
         \midrule
        LAVT \cite{lavt} (B) & 72.7 & 62.4 & 61.2 \\
        ReSTR \cite{rester} & 67.2 & 55.7 & 54.5 \\
        X-Decoder (T) \cite{xdecoder} & - & - & 61.9 \\
        X-Decoder (B) \cite{xdecoder} & - & - & 64.5 \\
        X-Decoder (L) \cite{xdecoder} & - & - & 64.6 \\
        \model{} (2D only) & 69.4 & 61.3 &  64.0 \\
        \model{} (2D-3D) & 69.2 & 61.3 & 64.1 \\
         \bottomrule
    \end{tabular}
    }
    \label{table:refcoco}
    \vspace{-0.8em}
\end{table}

\begin{table*}[ht!]
    \centering
    \caption{\textbf{Analysis of Box Head vs Mask Head} on ScanRefer Dataset with Acc@25 if not otherwise stated.}
    \label{table:box_mask}
    \begin{subtable}[t]{0.34\textwidth}
        \centering
        \caption{\textbf{Parametric vs Non-parametric Query}}
        \label{tab:query}
        \scriptsize %
        \setlength{\tabcolsep}{3pt} %
        \begin{tabular}{lcc}
            \toprule
            \textbf{Query Type} & \textbf{Box Head} & \textbf{Mask Head} \\
            \midrule
            Param & 23.9 & \textbf{54.4} \\
            Non-param & \textbf{34.5} & 43.9 \\ 
            \bottomrule
        \end{tabular}
    \end{subtable}
    \hfill
    \begin{subtable}[t]{0.31\textwidth}
        \centering
        \caption{\textbf{Updating Visual Features}}
        \label{tab:feat_attn}
        \scriptsize
        \setlength{\tabcolsep}{3pt}
        \begin{tabular}{lcc}
            \toprule
            \textbf{Feat Attn} & \textbf{Box Head} & \textbf{Mask Head} \\
            \midrule
            \ding{51} & 33.9 & \textbf{54.4} \\
            \ding{55} & \textbf{34.5} & 41.5 \\
            \bottomrule
        \end{tabular}
    \end{subtable}
    \hfill
    \begin{subtable}[t]{0.31\textwidth}
        \centering
        \caption{\textbf{Results at Various IoU Thresholds}}
        \label{tab:iou}
        \scriptsize
        \setlength{\tabcolsep}{3pt}
        \begin{tabular}{lcc}
            \toprule
            & \textbf{Acc@25} & \textbf{Acc@75} \\
            \midrule
            Box Head & 34.5 & 1.1 \\
            Mask Head & \textbf{54.4} & \textbf{33.2} \\
            \bottomrule
        \end{tabular}
    \end{subtable}
\end{table*}

\vspace{-3pt}
\begin{table}[ht!]
    \centering
    \caption{\textbf{Analysis of 2D training strategies} Acc@25 in \Det Setup}
    \vspace{2pt}
    \label{table:2d_3d_ablations}
    \resizebox{.49\textwidth}{!}{
    \begin{tabular}{p{5cm}cccc}
        \toprule
        \textbf{Model} & \textbf{Avg Accuracy} & \textbf{SR3D} & \textbf{NR3D} & \textbf{ScanRefer} \\
        \midrule
        \model{}-3D-only & 61.6 & 71.6 & 52.5 & 60.7 \\
        \model{}-2D-3D(w/o 2D-3D lifting)  & 62.7 & 73.0 & 53.5  & 61.5 \\
        \model{}-2D-3D(w/ 2D-3D lifting) & \textbf{64.3} & \textbf{73.0} & \textbf{56.3} & \textbf{63.5} \\
        \bottomrule
    \end{tabular}
    }
\end{table}

\begin{table}[ht!]
    \centering
    \caption{\textbf{Ablations} Acc@25 in \Det Setup}
    \label{table:ablations}
    \resizebox{.49\textwidth}{!}{
    \begin{tabular}{p{5cm}cccc}
        \toprule
        \textbf{Model} & \textbf{Avg Accuracy} & \textbf{SR3D} & \textbf{NR3D} & \textbf{ScanRefer} \\
        \midrule
        \model{} & \textbf{61.0} & \textbf{67.1} & \textbf{55.7} & \textbf{60.2} \\
        w/o mask decoder w/ box decoder & 39.3 & 38.9 & 33.2 & 45.7 \\
        w/o feature attn & 36.9 & 38.0 & 30.0 & 42.8 \\
        w/o pretrained 2D weights & 53.4 & 54.3 & 49.1 & 56.9 \\
        w/o mask bounding box loss & 56.8 & 64.3 & 49.5 & 56.7 \\
        \bottomrule
    \end{tabular}}
\end{table}

We also evaluate \model{} on the 2D Referential Grounding benchmarks \cite{refcoco} (\cref{table:refcoco}). We train two versions of our model: \model{} (2D only), which is trained exclusively on 2D datasets, and \model{} (2D-3D), which is trained on both 2D and 3D data. Our results show that co-training with 3D data does not degrade the performance of the version trained solely on 2D data. This demonstrates that it is indeed possible to train a single model for both 2D and 3D tasks. As we show in our experiments, this approach leads to significant improvements in 3D performance without negatively affecting 2D performance. In this work, due to our focus on improving 3D vision-language grounding and resource constraints, we did not train our model on additional 2D datasets, which is common in prior work. Scaling up these models with more 2D data and studying its impact on 3D vision-language grounding is a promising avenue for future research.

\vspace{-7pt}
\subsection{Ablations}
\label{ssec:ablations}

We ablate a series of design choices of our model  on referential grounding datasets of SR3D, NR3D, and ScanRefer on \cref{table:2d_3d_ablations} and \cref{table:ablations}; and on ScanRefer dataset in \cref{table:box_mask}. We have the following conclusions:

\textbf{1. Lifting 2D datasets to 3D improves 3D performance.} In \cref{table:2d_3d_ablations}, we compare three variants of our model: one trained only on 3D data, one trained with 3D data and 2D images without lifting them to 3D (where the 3D layers are skipped for 2D inputs, following~\cite{odin}), and our proposed approach of lifting 2D images to 3D pointmaps. We observe that incorporating 2D data improves performance in both scenarios, but our approach of lifting 2D images to 3D achieves the best results. In \cref{appendix:ai2thor}, we show in a more controlled setting that training without using 3D pointmaps—by skipping the 3D layers—results in significant overfitting to individual 2D-3D domains.

\textbf{2. Decoding  boxes is inferior to decoding segmentations.}
Shifting from decoding segmentation masks to decoding bounding boxes hurts performance (row 2 of \cref{table:ablations}), especially in tight IoU thresholds IoU@0.75, shown in \cref{tab:iou}. 

\textbf{3. Visual tokens updating through attending to language and queries during mask decoding is essential} for good performance in 3D referential grounding, as shown in row 3 of \cref{table:ablations}. 
This is potentially because the mask decoding head relies on dot-products of queries and features to predict masks; and thus having both object queries and visual features to be very well distinguished for different instances of the same object is crucial. This design choice is unique to mask decoding heads, as we show in \cref{tab:feat_attn}. Box-decoding models work similarly well irrespective of updating the visual tokens with language and object tokens. 
This variant is very close to ODIN's open vocabulary head, which also lacks such attention operations, and as we show it does not work well for referential language grounding. %

\textbf{4. 2D feature pretraining dramatically improves performance} as shown in row 4 of \cref{table:ablations}. 

\textbf{5. The predicted mask bounding box loss helps significantly} as shown in row 5 of \cref{table:ablations}.

\textbf{6.\;Non-Parametric Queries are crucial for decoding boxes prediction, while parametric queries work  well for decoding segments.}
 There are two popular choices for object queries: \textit{Parametric Queries} which are scene-independent learnable vectors, initialized from scratch, and are updated via attention. \textit{Non-Parametric Queries}, which are scene-dependent, and are typically initialized by doing Farthest Point Sampling on the input point clouds and encoding the corresponding xyz locations as query positional embeddings and corresponding features as query feature embeddings. Using non-parametric queries are crucial for box-decoding heads, while both queries work similarly well for mask-decoding heads (\cref{tab:query}).
 Box-decoding heads need to regress raw XYZ coordinates in 3D space; the search space is large and sparse---as most of it is empty---and parametric queries have difficulty handling such free space, as already mentioned in 3DETR \cite{3detr}.  Mask decoding uses dot-product between queries and visual tokens coming from the 3D backbone, and thus does not need to reason about 3D free space.

\subsection{2D-3D Generalization Test}
\label{appendix:ai2thor}
\begin{table}[ht]
    \centering
    \caption{\textbf{2D-3D Generalization Test}}
    \label{table:ai2thor}
    \resizebox{.49\textwidth}{!}{
    \begin{tabular}{p{5cm}cc}
        \toprule
        \textbf{Model} & \textbf{3D Sup. Classes} & \textbf{2D Sup. Classes} \\
        \midrule
        \model{} & 72.6 & 53.8 \\
        \model{} w/o 2D-to-3D lifting & 71.4 & 0.0  \\
        \rowcolor{gray!50}\model{} (Upper-Bound) & 69.7 & 84.2 \\
        \bottomrule
    \end{tabular}}
\end{table}

We study the generalization between 2D and 3D domains using the 3D Instance Segmentation benchmark in the AI2-THOR simulator \cite{ai2thor}. AI2-THOR consists of 120 object classes and provides posed RGB-D images of its 3D environments. We split these classes into two disjoint subsets: (a) 3D Supervised Classes (30) and (b) 2D Supervised Classes (90). We train \model{} in a 2D-3D setting where, for multi-view RGB-D inputs, the model is tasked with detecting objects from the 3D Supervised Classes subset. For single RGB image inputs, it is tasked with detecting objects from the 2D Supervised Classes subset. At test time, the model is evaluated in a multi-view posed RGB-D setup on all 120 object classes to assess its generalization across the two subsets. Results are presented in \cref{table:ai2thor}.

We find that when \model{} does not lift 2D images to 3D—simply skipping 3D attention layers in 2D batches, similar to ODIN~\cite{odin} (row 2 in \cref{table:ai2thor})—performance on 2D Supervised Classes drops to nearly zero. This suggests that without depth-based transformation, the model fails to learn from 2D supervision. However, when \model{} incorporates depth information, its performance on 2D Supervised Classes becomes non-trivial, indicating that it successfully learns from 2D supervision and applies this knowledge in the 3D domain at test time. In row 3, we report results for \model{} trained in 3D on all 120 classes, representing an upper bound on its performance given full 3D supervision. We observe that \model{} in row 1 remains significantly below the upper bound, highlighting the need for further improvements in 2D-to-3D generalization.

\subsection{Common failure modes of \model{}}
We identify three systematic failure modes in our model, illustrated in Figure-\ref{fig:failure} (see Appendix). 
    \\
    \textbf{Inclusion of distant outlier points in the predicted masks}: In the first image of Figure-\ref{fig:failure}, while \model{} accurately predicts the object, it also includes some distant points in the mask. This leads to a larger bounding box during the mask-to-bounding box conversion in post-processing, negatively affecting accuracy metrics. Although our proposed box loss mitigates this issue, it doesn't fully resolve it. 
    \\
    \textbf{Multiple instances of the same object being segmented together}: As shown in the middle image of Figure-\ref{fig:failure}, \model{} predicts both beds as a single output. Incorporating attention to language and queries and box loss helps reduce such errors, though they still persist. \\
    \textbf{Grounding failures} as seen in the third image of Figure-\ref{fig:failure}.     

The first two failure modes are specific to mask-decoding architectures, and similar issues have been noted by Mask3D \cite{mask3d} in their 3D instance segmentation tasks. Box-decoding architectures, on the other hand, generally avoid these problems. Nevertheless, we find that mask-decoding architectures offer significant advantages in other aspects, such as more accurate and fine-grained segmentation, making them valuable despite these challenges.

\vspace{-10pt}
\section{Conclusion}
\vspace{-2pt}
We present \model{}, a vision-language model that integrates 2D and 3D data to address data scarcity in 3D vision-language learning. By leveraging pre-trained 2D features, 2D-to-3D lifting strategies, and a novel mask decoder head, \model{} significantly outperforms prior methods in realistic embodied 3D vision settings while maintaining strong 2D understanding. Our extensive ablations validate key design choices: (1) Mask decoding is superior to box decoding, and each proposed component is crucial for its success. (2) Pre-trained 2D features improve performance, and co-training 3D vision-language tasks with 2D data provides additional gains. (3) Incorporating 2D-to-3D lifting strategies further enhances 3D understanding when training with 2D data. More broadly, our findings suggest that scaling 3D data is not the only path forward—leveraging 2D data and pre-trained features can effectively enhance 3D reasoning. We hope \model{} inspires further research into vision-language models that bridge 2D and 3D for real-world applications. 

While our current approach already achieves strong performance using significantly less 2D data than state-of-the-art 2D VLMs, scaling to larger 2D datasets is a promising next step to further improve 3D reasoning. Additionally, although \model{} is designed for static 3D scenes, extending it to dynamic 3D environments remains an important and exciting direction for future work. Finally, integrating \model{} with large-scale vision-language models holds strong potential for unifying 2D and 3D understanding at scale.

\ifthenelse{\boolean{icmlfinal}}{
\section*{Impact Statement}

Our work aims to advance 3D vision-language learning by leveraging pre-trained 2D models and 2D-to-3D lifting strategies to overcome the data scarcity in 3D datasets. This approach has the potential to make 3D vision more accessible and practical, particularly for embodied AI applications such as robotics, assistive technologies, and augmented reality. By reducing reliance on large-scale 3D annotations, our method could democratize research in 3D vision and lower computational and data collection costs.

However, as with any vision-language model, there are ethical considerations, including biases inherited from pre-trained 2D models, potential misuse in surveillance applications, and the environmental impact of large-scale model training. We encourage future work to explore methods for mitigating biases in pre-trained models and improving the transparency and interpretability of 3D vision-language systems.

Overall, we believe our contributions will positively impact the field of 3D vision-language learning, enabling progress toward more robust and generalizable embodied AI systems while being mindful of ethical considerations.

}{}
\section*{Acknowledgements}

The CMU authors were supported by an NSF Career award, ONR award N00014-23-1-2415, AFOSR Grant FA9550-23-1-0257, and DARPA No. HR00112490375 from the U.S. DARPA Friction for Accountability in Conversational Transactions (FACT) program. Ayush Jain is supported in part by the CMU Robotics Vision Fellowship. The authors express gratitude to Nikolaos Gkanatsios, Wen-Hsuan Chu, Gabriel Sarch, Mathew Bronars, and Naitik Khandelwal for their valuable feedback on the early draft of this work. Special thanks to Kyle Genova for several insightful discussions during the early stages of this project's development. The authors also thank Cortex team at Meta for valuable feedback and discussions.

\bibliography{bibtex,iclr2025_conference}

\begin{thebibliography}{74}
\providecommand{\natexlab}[1]{#1}
\providecommand{\url}[1]{\texttt{#1}}
\expandafter\ifx\csname urlstyle\endcsname\relax
  \providecommand{\doi}[1]{doi: #1}\else
  \providecommand{\doi}{doi: \begingroup \urlstyle{rm}\Url}\fi

\bibitem[Abdelrahman et~al.(2024)Abdelrahman, Ayman, Ahmed, Slim, and
  Elhoseiny]{abdelrahman2024cot3drefchainofthoughtsdataefficient3d}
Abdelrahman, E., Ayman, M., Ahmed, M., Slim, H., and Elhoseiny, M.
\newblock Cot3dref: Chain-of-thoughts data-efficient 3d visual grounding, 2024.
\newblock URL \url{https://arxiv.org/abs/2310.06214}.

\bibitem[Abdelreheem et~al.(2023)Abdelreheem, Olszewski, Lee, Wonka, and
  Achlioptas]{scanents}
Abdelreheem, A., Olszewski, K., Lee, H.-Y., Wonka, P., and Achlioptas, P.
\newblock Scanents3d: Exploiting phrase-to-3d-object correspondences for
  improved visio-linguistic models in 3d scenes, 2023.
\newblock URL \url{https://arxiv.org/abs/2212.06250}.

\bibitem[Achlioptas et~al.(2020)Achlioptas, Abdelreheem, Xia, Elhoseiny, and
  Guibas]{referit3d}
Achlioptas, P., Abdelreheem, A., Xia, F., Elhoseiny, M., and Guibas, L.
\newblock {ReferIt3D: Neural Listeners for Fine-Grained 3D Object
  Identification in Real-World Scenes}.
\newblock In \emph{Proc. ECCV}, 2020.

\bibitem[Ahn et~al.(2022)Ahn, Brohan, Brown, Chebotar, Cortes, David, Finn, Fu,
  Gopalakrishnan, Hausman, Herzog, Ho, Hsu, Ibarz, Ichter, Irpan, Jang, Ruano,
  Jeffrey, Jesmonth, Joshi, Julian, Kalashnikov, Kuang, Lee, Levine, Lu, Luu,
  Parada, Pastor, Quiambao, Rao, Rettinghouse, Reyes, Sermanet, Sievers, Tan,
  Toshev, Vanhoucke, Xia, Xiao, Xu, Xu, Yan, and Zeng]{saycan}
Ahn, M., Brohan, A., Brown, N., Chebotar, Y., Cortes, O., David, B., Finn, C.,
  Fu, C., Gopalakrishnan, K., Hausman, K., Herzog, A., Ho, D., Hsu, J., Ibarz,
  J., Ichter, B., Irpan, A., Jang, E., Ruano, R.~J., Jeffrey, K., Jesmonth, S.,
  Joshi, N.~J., Julian, R., Kalashnikov, D., Kuang, Y., Lee, K.-H., Levine, S.,
  Lu, Y., Luu, L., Parada, C., Pastor, P., Quiambao, J., Rao, K., Rettinghouse,
  J., Reyes, D., Sermanet, P., Sievers, N., Tan, C., Toshev, A., Vanhoucke, V.,
  Xia, F., Xiao, T., Xu, P., Xu, S., Yan, M., and Zeng, A.
\newblock Do as i can, not as i say: Grounding language in robotic affordances,
  2022.
\newblock URL \url{https://arxiv.org/abs/2204.01691}.

\bibitem[Arnaud et~al.(2025)Arnaud, McVay, Martin, Majumdar, Jatavallabhula,
  Thomas, Partsey, Dugas, Gejji, Sax, Berges, Henaff, Jain, Cao, Prasad,
  Kalakrishnan, Rabbat, Ballas, Assran, Maksymets, Rajeswaran, and
  Meier]{locate3d}
Arnaud, S., McVay, P., Martin, A., Majumdar, A., Jatavallabhula, K.~M., Thomas,
  P., Partsey, R., Dugas, D., Gejji, A., Sax, A., Berges, V.-P., Henaff, M.,
  Jain, A., Cao, A., Prasad, I., Kalakrishnan, M., Rabbat, M., Ballas, N.,
  Assran, M., Maksymets, O., Rajeswaran, A., and Meier, F.
\newblock Locate 3d: Real-world object localization via self-supervised
  learning in 3d, 2025.
\newblock URL \url{https://arxiv.org/abs/2504.14151}.

\bibitem[Azuma et~al.(2022)Azuma, Miyanishi, Kurita, and Kawanabe]{scanqa}
Azuma, D., Miyanishi, T., Kurita, S., and Kawanabe, M.
\newblock Scanqa: 3d question answering for spatial scene understanding.
\newblock In \emph{proceedings of the IEEE/CVF conference on computer vision
  and pattern recognition}, pp.\  19129--19139, 2022.

\bibitem[Baruch et~al.(2021)Baruch, Chen, Dehghan, Dimry, Feigin, Fu, Gebauer,
  Joffe, Kurz, Schwartz, et~al.]{arkitscenes}
Baruch, G., Chen, Z., Dehghan, A., Dimry, T., Feigin, Y., Fu, P., Gebauer, T.,
  Joffe, B., Kurz, D., Schwartz, A., et~al.
\newblock Arkitscenes: A diverse real-world dataset for 3d indoor scene
  understanding using mobile rgb-d data.
\newblock \emph{arXiv preprint arXiv:2111.08897}, 2021.

\bibitem[Carion et~al.(2020)Carion, Massa, Synnaeve, Usunier, Kirillov, and
  Zagoruyko]{Carion2020EndtoEndOD}
Carion, N., Massa, F., Synnaeve, G., Usunier, N., Kirillov, A., and Zagoruyko,
  S.
\newblock {End-to-End Object Detection with Transformers}.
\newblock In \emph{Proc. ECCV}, 2020.

\bibitem[Chang et~al.(2017)Chang, Dai, Funkhouser, Halber, Niessner, Savva,
  Song, Zeng, and Zhang]{matterport}
Chang, A., Dai, A., Funkhouser, T., Halber, M., Niessner, M., Savva, M., Song,
  S., Zeng, A., and Zhang, Y.
\newblock Matterport3d: Learning from rgb-d data in indoor environments.
\newblock \emph{arXiv preprint arXiv:1709.06158}, 2017.

\bibitem[Chang et~al.(2024)Chang, Wang, Pagani, and Stricker]{chang2024mikasa}
Chang, C.-P., Wang, S., Pagani, A., and Stricker, D.
\newblock Mikasa: Multi-key-anchor \& scene-aware transformer for 3d visual
  grounding.
\newblock In \emph{Proceedings of the IEEE/CVF Conference on Computer Vision
  and Pattern Recognition}, pp.\  14131--14140, 2024.

\bibitem[Chen et~al.(2024)Chen, Xu, Kirmani, Ichter, Driess, Florence, Sadigh,
  Guibas, and Xia]{spatialvlm}
Chen, B., Xu, Z., Kirmani, S., Ichter, B., Driess, D., Florence, P., Sadigh,
  D., Guibas, L., and Xia, F.
\newblock Spatialvlm: Endowing vision-language models with spatial reasoning
  capabilities, 2024.
\newblock URL \url{https://arxiv.org/abs/2401.12168}.

\bibitem[Chen et~al.(2020{\natexlab{a}})Chen, Chang, and
  Nie{\ss}ner]{scanrefer}
Chen, D.~Z., Chang, A., and Nie{\ss}ner, M.
\newblock {ScanRefer: 3D Object Localization in RGB-D Scans using Natural
  Language}.
\newblock In \emph{Proc. ECCV}, 2020{\natexlab{a}}.

\bibitem[Chen et~al.(2020{\natexlab{b}})Chen, Gholami, Nießner, and
  Chang]{scan2cap}
Chen, D.~Z., Gholami, A., Nießner, M., and Chang, A.~X.
\newblock Scan2cap: Context-aware dense captioning in rgb-d scans,
  2020{\natexlab{b}}.

\bibitem[Chen et~al.(2022)Chen, Guhur, Tapaswi, Schmid, and Laptev]{vil3d}
Chen, S., Guhur, P.-L., Tapaswi, M., Schmid, C., and Laptev, I.
\newblock Language conditioned spatial relation reasoning for 3d object
  grounding.
\newblock \emph{Advances in neural information processing systems},
  35:\penalty0 20522--20535, 2022.

\bibitem[Cheng et~al.(2024)Cheng, Yin, Fu, Guo, Yang, Kautz, Wang, and
  Liu]{spatialrgpt}
Cheng, A.-C., Yin, H., Fu, Y., Guo, Q., Yang, R., Kautz, J., Wang, X., and Liu,
  S.
\newblock Spatialrgpt: Grounded spatial reasoning in vision language models,
  2024.
\newblock URL \url{https://arxiv.org/abs/2406.01584}.

\bibitem[Cheng et~al.(2022)Cheng, Misra, Schwing, Kirillov, and Girdhar]{m2f}
Cheng, B., Misra, I., Schwing, A.~G., Kirillov, A., and Girdhar, R.
\newblock Masked-attention mask transformer for universal image segmentation.
\newblock 2022.

\bibitem[Chiang et~al.(2024)Chiang, Xu, Fu, Jacob, Zhang, Lee, Yu, Schenck,
  Rendleman, Shah, Xia, Hsu, Hoech, Florence, Kirmani, Singh, Sindhwani,
  Parada, Finn, Xu, Levine, and Tan]{mobilityvla}
Chiang, H.-T.~L., Xu, Z., Fu, Z., Jacob, M.~G., Zhang, T., Lee, T.-W.~E., Yu,
  W., Schenck, C., Rendleman, D., Shah, D., Xia, F., Hsu, J., Hoech, J.,
  Florence, P., Kirmani, S., Singh, S., Sindhwani, V., Parada, C., Finn, C.,
  Xu, P., Levine, S., and Tan, J.
\newblock Mobility vla: Multimodal instruction navigation with long-context
  vlms and topological graphs, 2024.
\newblock URL \url{https://arxiv.org/abs/2407.07775}.

\bibitem[Chibane et~al.(2022)Chibane, Engelmann, Anh~Tran, and
  Pons-Moll]{box2mask}
Chibane, J., Engelmann, F., Anh~Tran, T., and Pons-Moll, G.
\newblock Box2mask: Weakly supervised 3d semantic instance segmentation using
  bounding boxes.
\newblock In \emph{European conference on computer vision}, pp.\  681--699.
  Springer, 2022.

\bibitem[Dai et~al.(2017)Dai, Chang, Savva, Halber, Funkhouser, and
  Nie{\ss}ner]{scannet}
Dai, A., Chang, A.~X., Savva, M., Halber, M., Funkhouser, T., and Nie{\ss}ner,
  M.
\newblock Scannet: Richly-annotated 3d reconstructions of indoor scenes.
\newblock In \emph{Proceedings of the IEEE conference on computer vision and
  pattern recognition}, pp.\  5828--5839, 2017.

\bibitem[Deitke et~al.(2022)Deitke, Schwenk, Salvador, Weihs, Michel,
  VanderBilt, Schmidt, Ehsani, Kembhavi, and Farhadi]{objaverse}
Deitke, M., Schwenk, D., Salvador, J., Weihs, L., Michel, O., VanderBilt, E.,
  Schmidt, L., Ehsani, K., Kembhavi, A., and Farhadi, A.
\newblock Objaverse: A universe of annotated 3d objects.
\newblock \emph{arXiv preprint arXiv:2212.08051}, 2022.

\bibitem[Dubey et~al.(2024)Dubey, Jauhri, Pandey, Kadian, Al-Dahle, Letman,
  Mathur, Schelten, Yang, Fan, Goyal, Hartshorn, Yang, Mitra, Sravankumar,
  Korenev, Hinsvark, Rao, Zhang, Rodriguez, Gregerson, Spataru, tiste Roziere,
  Biron, Tang, Chern, Caucheteux, Nayak, Bi, Marra, McConnell, Keller, Touret,
  Wu, Wong, Ferrer, Nikolaidis, Allonsius, Song, Pintz, Livshits, Esiobu,
  Choudhary, Mahajan, Garcia-Olano, Perino, Hupkes, Lakomkin, AlBadawy,
  Lobanova, Dinan, Smith, Radenovic, Zhang, Synnaeve, Lee, Anderson, Nail,
  Mialon, Pang, Cucurell, Nguyen, Korevaar, Xu, Touvron, Zarov, Ibarra,
  Kloumann, Misra, Evtimov, Copet, Lee, Geffert, Vranes, Park, Mahadeokar,
  Shah, van~der Linde, Billock, Hong, Lee, Fu, Chi, Huang, Liu, Wang, Yu,
  Bitton, Spisak, Park, Rocca, Johnstun, Saxe, Jia, Alwala, Upasani, Plawiak,
  Li, neth Heafield, Stone, El-Arini, Iyer, Malik, ley Chiu, Bhalla,
  Rantala-Yeary, van~der Maaten, Chen, Tan, Jenkins, Martin, Madaan, Malo,
  Blecher, Landzaat, de~Oliveira, Muzzi, Pasupuleti, Singh, Paluri, Kardas,
  Oldham, Rita, Pavlova, Kambadur, Lewis, Si, Singh, Hassan, Goyal, Torabi,
  Bashlykov, Bogoychev, Chatterji, Duchenne, cCelebi, Alrassy, Zhang, Li,
  Vasi{\'c}, Weng, Bhargava, Dubal, Krishnan, Koura, Xu, He, Dong, Srinivasan,
  Ganapathy, Calderer, Cabral, Stojnic, Raileanu, Girdhar, Patel, main
  Sauvestre, Polidoro, Sumbaly, Taylor, Silva, Hou, Wang, Hosseini,
  Chennabasappa, Singh, Bell, Kim, Edunov, Nie, Narang, Raparthy, Shen, Wan,
  Bhosale, Zhang, Vandenhende, Batra, Whitman, Sootla, Collot, Gururangan,
  Borodinsky, Herman, Fowler, Sheasha, Georgiou, Scialom, Speckbacher,
  Mihaylov, Xiao, Karn, Goswami, Gupta, Ramanathan, Kerkez, Gonguet, Do,
  Vogeti, Petrovic, Chu, Xiong, Fu, ney Meers, Martinet, Wang, Tan, Xie, Jia,
  Wang, Goldschlag, Gaur, Babaei, Wen, Song, Zhang, Li, Mao, Coudert, Yan,
  Chen, Papakipos, Singh, Grattafiori, Jain, Kelsey, Shajnfeld, Gangidi,
  Victoria, Goldstand, Menon, Sharma, Boesenberg, Vaughan, Baevski, Feinstein,
  Kallet, Sangani, Yunus, Lupu, Alvarado, Caples, Gu, Ho, Poulton, Ryan,
  Ramchandani, Franco, Saraf, Chowdhury, Gabriel, Bharambe, Eisenman, Yazdan,
  James, Maurer, Leonhardi, Huang, Loyd, Paola, Paranjape, Liu, Wu, Ni,
  Hancock, Wasti, Spence, Stojkovic, Gamido, Montalvo, Parker, Burton, Mejia,
  Wang, Kim, Zhou, Hu, Chu, Cai, Tindal, Feichtenhofer, Civin, Beaty, Kreymer,
  Li, Wyatt, Adkins, Xu, Testuggine, David, Parikh, Liskovich, Foss, Wang, Le,
  Holland, Dowling, Jamil, Montgomery, Presani, Hahn, Wood, Brinkman, Arcaute,
  Dunbar, Smothers, Sun, Kreuk, Tian, Ozgenel, Caggioni, Guzm'an, Kanayet,
  Seide, Florez, Schwarz, Badeer, Swee, Halpern, Thattai, Herman, Sizov, Zhang,
  Lakshminarayanan, Shojanazeri, Zou, Wang, Zha, Habeeb, Rudolph, Suk,
  Aspegren, Goldman, Molybog, Tufanov, Veliche, Gat, Weissman, Geboski, Kohli,
  Asher, Gaya, Marcus, Tang, Chan, Zhen, Reizenstein, Teboul, Zhong, Jin, Yang,
  Cummings, Carvill, Shepard, McPhie, Torres, Ginsburg, Wang, Wu, KamHou,
  Saxena, Prasad, Khandelwal, Zand, Matosich, Veeraraghavan, Michelena, Li,
  Huang, Chawla, Lakhotia, Huang, Chen, Garg, Lavender, Silva, Bell, Zhang,
  Guo, Yu, Moshkovich, Wehrstedt, Khabsa, Avalani, Bhatt, Tsimpoukelli, Mankus,
  Hasson, Lennie, Reso, Groshev, Naumov, Lathi, Keneally, Seltzer, Valko,
  Restrepo, Patel, Vyatskov, Samvelyan, Clark, Macey, Wang, Hermoso, Metanat,
  Rastegari, Bansal, Santhanam, Parks, White, Bawa, Singhal, Egebo, Usunier,
  Laptev, Dong, Zhang, Cheng, Chernoguz, Hart, Salpekar, Kalinli, Kent, Parekh,
  Saab, Balaji, Rittner, Bontrager, Roux, Doll{\'a}r, Zvyagina, Ratanchandani,
  Yuvraj, Liang, Alao, Rodriguez, Ayub, Murthy, Nayani, Mitra, Li, Hogan,
  Battey, Wang, Maheswari, Howes, Rinott, Bondu, Datta, Chugh, Hunt, Dhillon,
  Sidorov, Pan, Verma, Yamamoto, Ramaswamy, Lindsay, Feng, Lin, Zha, Shankar,
  Zhang, Wang, Agarwal, Sajuyigbe, Chintala, Max, Chen, Kehoe, Satterfield,
  Govindaprasad, Gupta, Cho, Virk, Subramanian, Choudhury, Goldman, Remez,
  Glaser, Best, Kohler, Robinson, Li, Zhang, Matthews, Chou, Shaked,
  Vontimitta, Ajayi, Montanez, Mohan, Kumar, Mangla, Ionescu, Poenaru,
  Mihailescu, Ivanov, Li, Wang, Jiang, Bouaziz, Constable, Tang, Wang, Wu,
  Wang, Xia, Wu, Gao, Chen, Hu, Jia, Qi, Li, Zhang, Zhang, Adi, Nam, Wang, Hao,
  Qian, He, Rait, DeVito, Rosnbrick, Wen, Yang, and Zhao]{Dubey2024TheL3}
Dubey, A., Jauhri, A., Pandey, A., Kadian, A., Al-Dahle, A., Letman, A.,
  Mathur, A., Schelten, A., Yang, A., Fan, A., Goyal, A., Hartshorn, A.~S.,
  Yang, A., Mitra, A., Sravankumar, A., Korenev, A., Hinsvark, A., Rao, A.,
  Zhang, A., Rodriguez, A., Gregerson, A., Spataru, A., tiste Roziere, B.,
  Biron, B., Tang, B., Chern, B., Caucheteux, C., Nayak, C., Bi, C., Marra, C.,
  McConnell, C., Keller, C., Touret, C., Wu, C., Wong, C., Ferrer, C.~C.,
  Nikolaidis, C., Allonsius, D., Song, D., Pintz, D., Livshits, D., Esiobu, D.,
  Choudhary, D., Mahajan, D., Garcia-Olano, D., Perino, D., Hupkes, D.,
  Lakomkin, E., AlBadawy, E.~A., Lobanova, E., Dinan, E., Smith, E.~M.,
  Radenovic, F., Zhang, F., Synnaeve, G., Lee, G., Anderson, G.~L., Nail, G.,
  Mialon, G., Pang, G., Cucurell, G., Nguyen, H., Korevaar, H., Xu, H.,
  Touvron, H., Zarov, I., Ibarra, I.~A., Kloumann, I.~M., Misra, I., Evtimov,
  I., Copet, J., Lee, J., Geffert, J.~L., Vranes, J., Park, J., Mahadeokar, J.,
  Shah, J., van~der Linde, J., Billock, J., Hong, J., Lee, J., Fu, J., Chi, J.,
  Huang, J., Liu, J., Wang, J., Yu, J., Bitton, J., Spisak, J., Park, J.,
  Rocca, J., Johnstun, J., Saxe, J., Jia, J.-Q., Alwala, K.~V., Upasani, K.,
  Plawiak, K., Li, K., neth Heafield, K.-., Stone, K., El-Arini, K., Iyer, K.,
  Malik, K., ley Chiu, K., Bhalla, K., Rantala-Yeary, L., van~der Maaten, L.,
  Chen, L., Tan, L., Jenkins, L., Martin, L., Madaan, L., Malo, L., Blecher,
  L., Landzaat, L., de~Oliveira, L., Muzzi, M., Pasupuleti, M.~B., Singh, M.,
  Paluri, M., Kardas, M., Oldham, M., Rita, M., Pavlova, M., Kambadur, M.
  H.~M., Lewis, M., Si, M., Singh, M.~K., Hassan, M., Goyal, N., Torabi, N.,
  Bashlykov, N., Bogoychev, N., Chatterji, N.~S., Duchenne, O., cCelebi, O.,
  Alrassy, P., Zhang, P., Li, P., Vasi{\'c}, P., Weng, P., Bhargava, P., Dubal,
  P., Krishnan, P., Koura, P.~S., Xu, P., He, Q., Dong, Q., Srinivasan, R.,
  Ganapathy, R., Calderer, R., Cabral, R.~S., Stojnic, R., Raileanu, R.,
  Girdhar, R., Patel, R., main Sauvestre, R., Polidoro, R., Sumbaly, R.,
  Taylor, R., Silva, R., Hou, R., Wang, R., Hosseini, S., Chennabasappa, S.,
  Singh, S., Bell, S., Kim, S.~S., Edunov, S., Nie, S., Narang, S., Raparthy,
  S.~C., Shen, S., Wan, S., Bhosale, S., Zhang, S., Vandenhende, S., Batra, S.,
  Whitman, S., Sootla, S., Collot, S., Gururangan, S., Borodinsky, S., Herman,
  T., Fowler, T., Sheasha, T., Georgiou, T., Scialom, T., Speckbacher, T.,
  Mihaylov, T., Xiao, T., Karn, U., Goswami, V., Gupta, V., Ramanathan, V.,
  Kerkez, V., Gonguet, V., Do, V., Vogeti, V., Petrovic, V., Chu, W., Xiong,
  W., Fu, W., ney Meers, W., Martinet, X., Wang, X., Tan, X.~E., Xie, X., Jia,
  X., Wang, X., Goldschlag, Y., Gaur, Y., Babaei, Y., Wen, Y., Song, Y., Zhang,
  Y., Li, Y., Mao, Y., Coudert, Z.~D., Yan, Z., Chen, Z., Papakipos, Z., Singh,
  A.~K., Grattafiori, A., Jain, A., Kelsey, A., Shajnfeld, A., Gangidi, A.,
  Victoria, A., Goldstand, A., Menon, A., Sharma, A., Boesenberg, A., Vaughan,
  A., Baevski, A., Feinstein, A., Kallet, A., Sangani, A., Yunus, A., Lupu, A.,
  Alvarado, A., Caples, A., Gu, A., Ho, A., Poulton, A., Ryan, A., Ramchandani,
  A., Franco, A., Saraf, A., Chowdhury, A., Gabriel, A., Bharambe, A.,
  Eisenman, A., Yazdan, A., James, B., Maurer, B., Leonhardi, B., Huang,
  P.-Y.~B., Loyd, B., Paola, B.~D., Paranjape, B., Liu, B., Wu, B., Ni, B.,
  Hancock, B., Wasti, B., Spence, B., Stojkovic, B., Gamido, B., Montalvo, B.,
  Parker, C., Burton, C., Mejia, C., Wang, C., Kim, C., Zhou, C., Hu, C., Chu,
  C.-H., Cai, C., Tindal, C., Feichtenhofer, C., Civin, D., Beaty, D., Kreymer,
  D., Li, S.-W., Wyatt, D., Adkins, D., Xu, D., Testuggine, D., David, D.,
  Parikh, D., Liskovich, D., Foss, D., Wang, D., Le, D., Holland, D., Dowling,
  E., Jamil, E., Montgomery, E., Presani, E., Hahn, E., Wood, E., Brinkman, E.,
  Arcaute, E., Dunbar, E., Smothers, E., Sun, F., Kreuk, F., Tian, F., Ozgenel,
  F., Caggioni, F., Guzm'an, F., Kanayet, F.~J., Seide, F., Florez, G.~M.,
  Schwarz, G., Badeer, G., Swee, G., Halpern, G., Thattai, G., Herman, G.,
  Sizov, G.~G., Zhang, G., Lakshminarayanan, G., Shojanazeri, H., Zou, H.,
  Wang, H., Zha, H., Habeeb, H., Rudolph, H., Suk, H., Aspegren, H., Goldman,
  H., Molybog, I., Tufanov, I., Veliche, I.-E., Gat, I., Weissman, J., Geboski,
  J., Kohli, J., Asher, J., Gaya, J.-B., Marcus, J., Tang, J., Chan, J., Zhen,
  J., Reizenstein, J., Teboul, J., Zhong, J., Jin, J., Yang, J., Cummings, J.,
  Carvill, J., Shepard, J., McPhie, J., Torres, J., Ginsburg, J., Wang, J., Wu,
  K., KamHou, U., Saxena, K., Prasad, K., Khandelwal, K., Zand, K., Matosich,
  K., Veeraraghavan, K., Michelena, K., Li, K., Huang, K., Chawla, K.,
  Lakhotia, K., Huang, K., Chen, L., Garg, L., Lavender, A., Silva, L., Bell,
  L., Zhang, L., Guo, L., Yu, L., Moshkovich, L., Wehrstedt, L., Khabsa, M.,
  Avalani, M., Bhatt, M., Tsimpoukelli, M., Mankus, M., Hasson, M., Lennie, M.,
  Reso, M., Groshev, M., Naumov, M., Lathi, M., Keneally, M., Seltzer, M.~L.,
  Valko, M., Restrepo, M., Patel, M., Vyatskov, M., Samvelyan, M., Clark, M.,
  Macey, M., Wang, M., Hermoso, M.~J., Metanat, M., Rastegari, M., Bansal, M.,
  Santhanam, N., Parks, N., White, N., Bawa, N., Singhal, N., Egebo, N.,
  Usunier, N., Laptev, N.~P., Dong, N., Zhang, N., Cheng, N., Chernoguz, O.,
  Hart, O., Salpekar, O., Kalinli, O., Kent, P., Parekh, P., Saab, P., Balaji,
  P., Rittner, P., Bontrager, P., Roux, P., Doll{\'a}r, P., Zvyagina, P.,
  Ratanchandani, P., Yuvraj, P., Liang, Q., Alao, R., Rodriguez, R., Ayub, R.,
  Murthy, R., Nayani, R., Mitra, R., Li, R., Hogan, R., Battey, R., Wang, R.,
  Maheswari, R., Howes, R., Rinott, R., Bondu, S.~J., Datta, S., Chugh, S.,
  Hunt, S., Dhillon, S., Sidorov, S., Pan, S., Verma, S., Yamamoto, S.,
  Ramaswamy, S., Lindsay, S., Feng, S., Lin, S., Zha, S.~C., Shankar, S.,
  Zhang, S., Wang, S., Agarwal, S., Sajuyigbe, S., Chintala, S., Max, S., Chen,
  S., Kehoe, S., Satterfield, S., Govindaprasad, S., Gupta, S., Cho, S.-B.,
  Virk, S., Subramanian, S., Choudhury, S., Goldman, S., Remez, T., Glaser, T.,
  Best, T., Kohler, T., Robinson, T., Li, T., Zhang, T., Matthews, T., Chou,
  T., Shaked, T., Vontimitta, V., Ajayi, V., Montanez, V., Mohan, V., Kumar,
  V.~S., Mangla, V., Ionescu, V., Poenaru, V.~A., Mihailescu, V.~T., Ivanov,
  V., Li, W., Wang, W., Jiang, W., Bouaziz, W., Constable, W., Tang, X., Wang,
  X., Wu, X., Wang, X., Xia, X., Wu, X., Gao, X., Chen, Y., Hu, Y., Jia, Y.,
  Qi, Y., Li, Y., Zhang, Y., Zhang, Y., Adi, Y., Nam, Y., Wang, Y., Hao, Y.,
  Qian, Y., He, Y., Rait, Z., DeVito, Z., Rosnbrick, Z., Wen, Z., Yang, Z., and
  Zhao, Z.
\newblock The llama 3 herd of models.
\newblock \emph{ArXiv}, abs/2407.21783, 2024.
\newblock URL \url{https://api.semanticscholar.org/CorpusID:271571434}.

\bibitem[Fang et~al.(2021)Fang, Jain, Sarch, Harley, and Fragkiadaki]{move}
Fang, Z., Jain, A., Sarch, G., Harley, A.~W., and Fragkiadaki, K.
\newblock Move to see better: Self-improving embodied object detection, 2021.
\newblock URL \url{https://arxiv.org/abs/2012.00057}.

\bibitem[Hong et~al.(2023)Hong, Zhen, Chen, Zheng, Du, Chen, and Gan]{3dllm}
Hong, Y., Zhen, H., Chen, P., Zheng, S., Du, Y., Chen, Z., and Gan, C.
\newblock 3d-llm: Injecting the 3d world into large language models.
\newblock \emph{Advances in Neural Information Processing Systems},
  36:\penalty0 20482--20494, 2023.

\bibitem[Hu et~al.(2021)Hu, Shen, Wallis, Allen-Zhu, Li, Wang, Wang, and
  Chen]{hu2021lora}
Hu, E.~J., Shen, Y., Wallis, P., Allen-Zhu, Z., Li, Y., Wang, S., Wang, L., and
  Chen, W.
\newblock Lora: Low-rank adaptation of large language models.
\newblock \emph{arXiv preprint arXiv:2106.09685}, 2021.

\bibitem[Jain et~al.(2022)Jain, Gkanatsios, Mediratta, and Fragkiadaki]{butd}
Jain, A., Gkanatsios, N., Mediratta, I., and Fragkiadaki, K.
\newblock Bottom up top down detection transformers for language grounding in
  images and point clouds.
\newblock In \emph{European Conference on Computer Vision}, pp.\  417--433.
  Springer, 2022.

\bibitem[Jain et~al.(2024)Jain, Katara, Gkanatsios, Harley, Sarch, Aggarwal,
  Chaudhary, and Fragkiadaki]{odin}
Jain, A., Katara, P., Gkanatsios, N., Harley, A.~W., Sarch, G., Aggarwal, K.,
  Chaudhary, V., and Fragkiadaki, K.
\newblock Odin: A single model for 2d and 3d segmentation.
\newblock In \emph{Proceedings of the IEEE/CVF Conference on Computer Vision
  and Pattern Recognition}, pp.\  3564--3574, 2024.

\bibitem[Kamath et~al.(2021)Kamath, Singh, LeCun, Misra, Synnaeve, and
  Carion]{Kamath2021MDETRM}
Kamath, A., Singh, M., LeCun, Y.~A., Misra, I., Synnaeve, G., and Carion, N.
\newblock {MDETR - Modulated Detection for End-to-End Multi-Modal
  Understanding}.
\newblock In \emph{Proc. ICCV}, 2021.

\bibitem[Kazemzadeh et~al.(2014)Kazemzadeh, Ordonez, Matten, and Berg]{refcoco}
Kazemzadeh, S., Ordonez, V., Matten, M., and Berg, T.
\newblock {R}efer{I}t{G}ame: Referring to objects in photographs of natural
  scenes.
\newblock In Moschitti, A., Pang, B., and Daelemans, W. (eds.),
  \emph{Proceedings of the 2014 Conference on Empirical Methods in Natural
  Language Processing ({EMNLP})}, pp.\  787--798, Doha, Qatar, October 2014.
  Association for Computational Linguistics.
\newblock \doi{10.3115/v1/D14-1086}.
\newblock URL \url{https://aclanthology.org/D14-1086/}.

\bibitem[Kim et~al.(2022)Kim, Kim, Lan, Zeng, and Kwak]{rester}
Kim, N., Kim, D., Lan, C., Zeng, W., and Kwak, S.
\newblock Restr: Convolution-free referring image segmentation using
  transformers, 2022.
\newblock URL \url{https://arxiv.org/abs/2203.16768}.

\bibitem[Kirillov et~al.(2023)Kirillov, Mintun, Ravi, Mao, Rolland, Gustafson,
  Xiao, Whitehead, Berg, Lo, et~al.]{sam}
Kirillov, A., Mintun, E., Ravi, N., Mao, H., Rolland, C., Gustafson, L., Xiao,
  T., Whitehead, S., Berg, A.~C., Lo, W.-Y., et~al.
\newblock Segment anything.
\newblock \emph{arXiv preprint arXiv:2304.02643}, 2023.

\bibitem[Kolve et~al.(2017)Kolve, Mottaghi, Han, VanderBilt, Weihs, Herrasti,
  Deitke, Ehsani, Gordon, Zhu, et~al.]{ai2thor}
Kolve, E., Mottaghi, R., Han, W., VanderBilt, E., Weihs, L., Herrasti, A.,
  Deitke, M., Ehsani, K., Gordon, D., Zhu, Y., et~al.
\newblock Ai2-thor: An interactive 3d environment for visual ai.
\newblock \emph{arXiv preprint arXiv:1712.05474}, 2017.

\bibitem[Koukounas et~al.(2024)Koukounas, Mastrapas, G{\"u}nther, Wang,
  Martens, Mohr, Sturua, Akram, Mart{\'\i}nez, Ognawala,
  et~al.]{koukounas2024jina}
Koukounas, A., Mastrapas, G., G{\"u}nther, M., Wang, B., Martens, S., Mohr, I.,
  Sturua, S., Akram, M.~K., Mart{\'\i}nez, J.~F., Ognawala, S., et~al.
\newblock Jina clip: Your clip model is also your text retriever.
\newblock \emph{arXiv preprint arXiv:2405.20204}, 2024.

\bibitem[Kundu et~al.(2020{\natexlab{a}})Kundu, Yin, Fathi, Ross, Brewington,
  Funkhouser, and Pantofaru]{kundu2020virtualmultiviewfusion3d}
Kundu, A., Yin, X., Fathi, A., Ross, D., Brewington, B., Funkhouser, T., and
  Pantofaru, C.
\newblock Virtual multi-view fusion for 3d semantic segmentation,
  2020{\natexlab{a}}.
\newblock URL \url{https://arxiv.org/abs/2007.13138}.

\bibitem[Kundu et~al.(2020{\natexlab{b}})Kundu, Yin, Fathi, Ross, Brewington,
  Funkhouser, and Pantofaru]{virtualview}
Kundu, A., Yin, X., Fathi, A., Ross, D., Brewington, B., Funkhouser, T., and
  Pantofaru, C.
\newblock Virtual multi-view fusion for 3d semantic segmentation.
\newblock In \emph{Computer Vision--ECCV 2020: 16th European Conference,
  Glasgow, UK, August 23--28, 2020, Proceedings, Part XXIV 16}, pp.\  518--535.
  Springer, 2020{\natexlab{b}}.

\bibitem[Lai et~al.(2023)Lai, Yuan, Chu, Chen, Hu, and Jia]{maft}
Lai, X., Yuan, Y., Chu, R., Chen, Y., Hu, H., and Jia, J.
\newblock Mask-attention-free transformer for 3d instance segmentation.
\newblock In \emph{Proceedings of the IEEE/CVF International Conference on
  Computer Vision}, pp.\  3693--3703, 2023.

\bibitem[Li et~al.(2023)Li, Li, Savarese, and Hoi]{blip2}
Li, J., Li, D., Savarese, S., and Hoi, S.
\newblock Blip-2: Bootstrapping language-image pre-training with frozen image
  encoders and large language models.
\newblock In \emph{International conference on machine learning}, pp.\
  19730--19742. PMLR, 2023.

\bibitem[Li et~al.(2022)Li, Zhang, Zhang, Yang, Li, Zhong, Wang, Yuan, Zhang,
  Hwang, et~al.]{glip}
Li, L.~H., Zhang, P., Zhang, H., Yang, J., Li, C., Zhong, Y., Wang, L., Yuan,
  L., Zhang, L., Hwang, J.-N., et~al.
\newblock Grounded language-image pre-training.
\newblock In \emph{Proceedings of the IEEE/CVF Conference on Computer Vision
  and Pattern Recognition}, pp.\  10965--10975, 2022.

\bibitem[Lin et~al.(2014)Lin, Maire, Belongie, Hays, Perona, Ramanan,
  Doll{\'a}r, and Zitnick]{coco}
Lin, T.-Y., Maire, M., Belongie, S., Hays, J., Perona, P., Ramanan, D.,
  Doll{\'a}r, P., and Zitnick, C.~L.
\newblock Microsoft coco: Common objects in context.
\newblock In \emph{Computer Vision--ECCV 2014: 13th European Conference,
  Zurich, Switzerland, September 6-12, 2014, Proceedings, Part V 13}, pp.\
  740--755. Springer, 2014.

\bibitem[Liu et~al.(2021)Liu, Lin, Cao, Hu, Wei, Zhang, Lin, and Guo]{swin}
Liu, Z., Lin, Y., Cao, Y., Hu, H., Wei, Y., Zhang, Z., Lin, S., and Guo, B.
\newblock Swin transformer: Hierarchical vision transformer using shifted
  windows.
\newblock In \emph{Proceedings of the IEEE/CVF international conference on
  computer vision}, pp.\  10012--10022, 2021.

\bibitem[Lu et~al.(2023)Lu, Deng, Wang, He, and Zhang]{queryformer}
Lu, J., Deng, J., Wang, C., He, J., and Zhang, T.
\newblock Query refinement transformer for 3d instance segmentation.
\newblock In \emph{Proceedings of the IEEE/CVF International Conference on
  Computer Vision}, pp.\  18516--18526, 2023.

\bibitem[Luo et~al.(2022)Luo, Fu, Kong, Gao, Ren, Shen, Xia, and Liu]{3dsps}
Luo, J., Fu, J., Kong, X., Gao, C., Ren, H., Shen, H., Xia, H., and Liu, S.
\newblock 3d-sps: Single-stage 3d visual grounding via referred point
  progressive selection.
\newblock In \emph{Proceedings of the IEEE/CVF Conference on Computer Vision
  and Pattern Recognition}, pp.\  16454--16463, 2022.

\bibitem[Ma et~al.(2022)Ma, Yong, Zheng, Li, Liang, Zhu, and Huang]{sqa3d}
Ma, X., Yong, S., Zheng, Z., Li, Q., Liang, Y., Zhu, S.-C., and Huang, S.
\newblock Sqa3d: Situated question answering in 3d scenes.
\newblock \emph{arXiv preprint arXiv:2210.07474}, 2022.

\bibitem[Misra et~al.(2021)Misra, Girdhar, and Joulin]{3detr}
Misra, I., Girdhar, R., and Joulin, A.
\newblock An end-to-end transformer model for 3d object detection.
\newblock In \emph{Proceedings of the IEEE/CVF international conference on
  computer vision}, pp.\  2906--2917, 2021.

\bibitem[Oquab et~al.(2024)Oquab, Darcet, Moutakanni, Vo, Szafraniec, Khalidov,
  Fernandez, Haziza, Massa, El-Nouby, Assran, Ballas, Galuba, Howes, Huang, Li,
  Misra, Rabbat, Sharma, Synnaeve, Xu, Jegou, Mairal, Labatut, Joulin, and
  Bojanowski]{oquab2024dinov2learningrobustvisual}
Oquab, M., Darcet, T., Moutakanni, T., Vo, H., Szafraniec, M., Khalidov, V.,
  Fernandez, P., Haziza, D., Massa, F., El-Nouby, A., Assran, M., Ballas, N.,
  Galuba, W., Howes, R., Huang, P.-Y., Li, S.-W., Misra, I., Rabbat, M.,
  Sharma, V., Synnaeve, G., Xu, H., Jegou, H., Mairal, J., Labatut, P., Joulin,
  A., and Bojanowski, P.
\newblock Dinov2: Learning robust visual features without supervision, 2024.
\newblock URL \url{https://arxiv.org/abs/2304.07193}.

\bibitem[Peng et~al.(2023{\natexlab{a}})Peng, Li, He, Galley, and Gao]{vicuna}
Peng, B., Li, C., He, P., Galley, M., and Gao, J.
\newblock Instruction tuning with gpt-4.
\newblock \emph{arXiv preprint arXiv:2304.03277}, 2023{\natexlab{a}}.

\bibitem[Peng et~al.(2023{\natexlab{b}})Peng, Genova, Jiang, Tagliasacchi,
  Pollefeys, Funkhouser, et~al.]{openscene}
Peng, S., Genova, K., Jiang, C., Tagliasacchi, A., Pollefeys, M., Funkhouser,
  T., et~al.
\newblock Openscene: 3d scene understanding with open vocabularies.
\newblock In \emph{Proceedings of the IEEE/CVF Conference on Computer Vision
  and Pattern Recognition}, pp.\  815--824, 2023{\natexlab{b}}.

\bibitem[Radford et~al.(2021)Radford, Kim, Hallacy, Ramesh, Goh, Agarwal,
  Sastry, Askell, Mishkin, Clark, et~al.]{clip}
Radford, A., Kim, J.~W., Hallacy, C., Ramesh, A., Goh, G., Agarwal, S., Sastry,
  G., Askell, A., Mishkin, P., Clark, J., et~al.
\newblock Learning transferable visual models from natural language
  supervision.
\newblock In \emph{International conference on machine learning}, pp.\
  8748--8763. PMLR, 2021.

\bibitem[Raffel et~al.(2020)Raffel, Shazeer, Roberts, Lee, Narang, Matena,
  Zhou, Li, and Liu]{raffel2020exploring}
Raffel, C., Shazeer, N., Roberts, A., Lee, K., Narang, S., Matena, M., Zhou,
  Y., Li, W., and Liu, P.~J.
\newblock Exploring the limits of transfer learning with a unified text-to-text
  transformer.
\newblock \emph{Journal of machine learning research}, 21\penalty0
  (140):\penalty0 1--67, 2020.

\bibitem[Rezatofighi et~al.(2019)Rezatofighi, Tsoi, Gwak, Sadeghian, Reid, and
  Savarese]{rezatofighi2019generalized}
Rezatofighi, H., Tsoi, N., Gwak, J., Sadeghian, A., Reid, I., and Savarese, S.
\newblock Generalized intersection over union: A metric and a loss for bounding
  box regression.
\newblock In \emph{Proceedings of the IEEE/CVF conference on computer vision
  and pattern recognition}, pp.\  658--666, 2019.

\bibitem[Robert et~al.(2022)Robert, Vallet, and Landrieu]{deepview}
Robert, D., Vallet, B., and Landrieu, L.
\newblock Learning multi-view aggregation in the wild for large-scale 3d
  semantic segmentation.
\newblock In \emph{Proceedings of the IEEE/CVF Conference on Computer Vision
  and Pattern Recognition}, pp.\  5575--5584, 2022.

\bibitem[Roh et~al.(2022)Roh, Desingh, Farhadi, and Fox]{roh2022languagerefer}
Roh, J., Desingh, K., Farhadi, A., and Fox, D.
\newblock Languagerefer: Spatial-language model for 3d visual grounding.
\newblock In \emph{Conference on Robot Learning}, pp.\  1046--1056. PMLR, 2022.

\bibitem[Rozenberszki et~al.(2022)Rozenberszki, Litany, and Dai]{scannet200}
Rozenberszki, D., Litany, O., and Dai, A.
\newblock Language-grounded indoor 3d semantic segmentation in the wild.
\newblock In \emph{European Conference on Computer Vision}, pp.\  125--141.
  Springer, 2022.

\bibitem[Rukhovich et~al.(2022)Rukhovich, Vorontsova, and Konushin]{imvoxelnet}
Rukhovich, D., Vorontsova, A., and Konushin, A.
\newblock Imvoxelnet: Image to voxels projection for monocular and multi-view
  general-purpose 3d object detection.
\newblock In \emph{Proceedings of the IEEE/CVF Winter Conference on
  Applications of Computer Vision}, pp.\  2397--2406, 2022.

\bibitem[Schult et~al.(2023)Schult, Engelmann, Hermans, Litany, Tang, and
  Leibe]{mask3d}
Schult, J., Engelmann, F., Hermans, A., Litany, O., Tang, S., and Leibe, B.
\newblock Mask3d: Mask transformer for 3d semantic instance segmentation.
\newblock In \emph{2023 IEEE International Conference on Robotics and
  Automation (ICRA)}, pp.\  8216--8223. IEEE, 2023.

\bibitem[Siddiqui et~al.(2023)Siddiqui, Porzi, Bul{\`o}, M{\"u}ller,
  Nie{\ss}ner, Dai, and Kontschieder]{pl}
Siddiqui, Y., Porzi, L., Bul{\`o}, S.~R., M{\"u}ller, N., Nie{\ss}ner, M., Dai,
  A., and Kontschieder, P.
\newblock Panoptic lifting for 3d scene understanding with neural fields.
\newblock In \emph{Proceedings of the IEEE/CVF Conference on Computer Vision
  and Pattern Recognition}, pp.\  9043--9052, 2023.

\bibitem[Straub et~al.(2024)Straub, DeTone, Shen, Yang, Sweeney, and
  Newcombe]{straub2024efm3dbenchmarkmeasuringprogress}
Straub, J., DeTone, D., Shen, T., Yang, N., Sweeney, C., and Newcombe, R.
\newblock Efm3d: A benchmark for measuring progress towards 3d egocentric
  foundation models, 2024.
\newblock URL \url{https://arxiv.org/abs/2406.10224}.

\bibitem[Sturm et~al.(2012)Sturm, Engelhard, Endres, Burgard, and
  Cremers]{6385773}
Sturm, J., Engelhard, N., Endres, F., Burgard, W., and Cremers, D.
\newblock A benchmark for the evaluation of rgb-d slam systems.
\newblock In \emph{2012 IEEE/RSJ International Conference on Intelligent Robots
  and Systems}, pp.\  573--580, 2012.
\newblock \doi{10.1109/IROS.2012.6385773}.

\bibitem[Wald et~al.(2019)Wald, Avetisyan, Navab, Tombari, and
  Nie{\ss}ner]{3rscan}
Wald, J., Avetisyan, A., Navab, N., Tombari, F., and Nie{\ss}ner, M.
\newblock Rio: 3d object instance re-localization in changing indoor
  environments.
\newblock In \emph{Proceedings of the IEEE/CVF International Conference on
  Computer Vision}, pp.\  7658--7667, 2019.

\bibitem[Wang et~al.(2024)Wang, Xu, Dai, Xiang, Deng, Tong, and Yang]{moge}
Wang, R., Xu, S., Dai, C., Xiang, J., Deng, Y., Tong, X., and Yang, J.
\newblock Moge: Unlocking accurate monocular geometry estimation for
  open-domain images with optimal training supervision.
\newblock \emph{arXiv preprint arXiv:2410.19115}, 2024.

\bibitem[Wang et~al.(2023)Wang, Mao, Zhu, Xu, Lyu, Li, Chen, Zhang, Chen, Xue,
  Liu, Lu, Lin, and Pang]{embscan}
Wang, T., Mao, X., Zhu, C., Xu, R., Lyu, R., Li, P., Chen, X., Zhang, W., Chen,
  K., Xue, T., Liu, X., Lu, C., Lin, D., and Pang, J.
\newblock Embodiedscan: A holistic multi-modal 3d perception suite towards
  embodied ai, 2023.
\newblock URL \url{https://arxiv.org/abs/2312.16170}.

\bibitem[Yadav et~al.(2023)Yadav, Ramrakhya, Ramakrishnan, Gervet, Turner,
  Gokaslan, Maestre, Chang, Batra, Savva, et~al.]{hm3d}
Yadav, K., Ramrakhya, R., Ramakrishnan, S.~K., Gervet, T., Turner, J.,
  Gokaslan, A., Maestre, N., Chang, A.~X., Batra, D., Savva, M., et~al.
\newblock Habitat-matterport 3d semantics dataset.
\newblock In \emph{Proceedings of the IEEE/CVF Conference on Computer Vision
  and Pattern Recognition}, pp.\  4927--4936, 2023.

\bibitem[Yang et~al.(2021{\natexlab{a}})Yang, Zhang, Wang, and
  Luo]{Yang2021SAT2S}
Yang, Z., Zhang, S., Wang, L., and Luo, J.
\newblock {SAT: 2D Semantics Assisted Training for 3D Visual Grounding}.
\newblock In \emph{Proc. ICCV}, 2021{\natexlab{a}}.

\bibitem[Yang et~al.(2021{\natexlab{b}})Yang, Zhang, Wang, and
  Luo]{yang2021sat}
Yang, Z., Zhang, S., Wang, L., and Luo, J.
\newblock Sat: 2d semantics assisted training for 3d visual grounding.
\newblock In \emph{Proceedings of the IEEE/CVF International Conference on
  Computer Vision}, pp.\  1856--1866, 2021{\natexlab{b}}.

\bibitem[Yang et~al.(2022)Yang, Wang, Tang, Chen, Zhao, and Torr]{lavt}
Yang, Z., Wang, J., Tang, Y., Chen, K., Zhao, H., and Torr, P. H.~S.
\newblock Lavt: Language-aware vision transformer for referring image
  segmentation, 2022.
\newblock URL \url{https://arxiv.org/abs/2112.02244}.

\bibitem[Yeshwanth et~al.(2023)Yeshwanth, Liu, Nie{\ss}ner, and Dai]{scannetpp}
Yeshwanth, C., Liu, Y.-C., Nie{\ss}ner, M., and Dai, A.
\newblock Scannet++: A high-fidelity dataset of 3d indoor scenes.
\newblock In \emph{Proceedings of the IEEE/CVF International Conference on
  Computer Vision}, pp.\  12--22, 2023.

\bibitem[Yuan et~al.(2021{\natexlab{a}})Yuan, Yan, Liao, Zhang, Li, and
  Cui]{Yuan2021InstanceReferCH}
Yuan, Z., Yan, X., Liao, Y., Zhang, R., Li, Z., and Cui, S.
\newblock {InstanceRefer: Cooperative Holistic Understanding for Visual
  Grounding on Point Clouds through Instance Multi-level Contextual Referring}.
\newblock In \emph{Proc. ICCV}, 2021{\natexlab{a}}.

\bibitem[Yuan et~al.(2021{\natexlab{b}})Yuan, Yan, Liao, Zhang, Wang, Li, and
  Cui]{yuan2021instancerefer}
Yuan, Z., Yan, X., Liao, Y., Zhang, R., Wang, S., Li, Z., and Cui, S.
\newblock Instancerefer: Cooperative holistic understanding for visual
  grounding on point clouds through instance multi-level contextual referring.
\newblock In \emph{Proceedings of the IEEE/CVF International Conference on
  Computer Vision}, pp.\  1791--1800, 2021{\natexlab{b}}.

\bibitem[Zhang et~al.(2023)Zhang, Gong, and Chang]{multi3drefer}
Zhang, Y., Gong, Z., and Chang, A.~X.
\newblock Multi3drefer: Grounding text description to multiple 3d objects.
\newblock In \emph{Proceedings of the IEEE/CVF International Conference on
  Computer Vision}, pp.\  15225--15236, 2023.

\bibitem[Zheng et~al.(2024)Zheng, Huang, Zhao, Zhong, and Wang]{navillm}
Zheng, D., Huang, S., Zhao, L., Zhong, Y., and Wang, L.
\newblock Towards learning a generalist model for embodied navigation.
\newblock In \emph{Proceedings of the IEEE/CVF Conference on Computer Vision
  and Pattern Recognition}, pp.\  13624--13634, 2024.

\bibitem[Zhu et~al.(2023{\natexlab{a}})Zhu, Zhang, Wang, Liu, and
  Chen]{object2scene}
Zhu, C., Zhang, W., Wang, T., Liu, X., and Chen, K.
\newblock Object2scene: Putting objects in context for open-vocabulary 3d
  detection.
\newblock \emph{arXiv preprint arXiv:2309.09456}, 2023{\natexlab{a}}.

\bibitem[Zhu et~al.(2024{\natexlab{a}})Zhu, Wang, Zhang, Pang, and Liu]{l3d}
Zhu, C., Wang, T., Zhang, W., Pang, J., and Liu, X.
\newblock Llava-3d: A simple yet effective pathway to empowering lmms with
  3d-awareness, 2024{\natexlab{a}}.
\newblock URL \url{https://arxiv.org/abs/2409.18125}.

\bibitem[Zhu et~al.(2023{\natexlab{b}})Zhu, Ma, Chen, Deng, Huang, and
  Li]{3dvista}
Zhu, Z., Ma, X., Chen, Y., Deng, Z., Huang, S., and Li, Q.
\newblock 3d-vista: Pre-trained transformer for 3d vision and text alignment.
\newblock In \emph{Proceedings of the IEEE/CVF International Conference on
  Computer Vision}, pp.\  2911--2921, 2023{\natexlab{b}}.

\bibitem[Zhu et~al.(2024{\natexlab{b}})Zhu, Zhang, Ma, Niu, Chen, Jia, Deng,
  Huang, and Li]{pq3d}
Zhu, Z., Zhang, Z., Ma, X., Niu, X., Chen, Y., Jia, B., Deng, Z., Huang, S.,
  and Li, Q.
\newblock Unifying 3d vision-language understanding via promptable queries.
\newblock \emph{arXiv preprint arXiv:2405.11442}, 2024{\natexlab{b}}.

\bibitem[Zou et~al.(2023)Zou, Dou, Yang, Gan, Li, Li, Dai, Behl, Wang, Yuan,
  et~al.]{xdecoder}
Zou, X., Dou, Z.-Y., Yang, J., Gan, Z., Li, L., Li, C., Dai, X., Behl, H.,
  Wang, J., Yuan, L., et~al.
\newblock Generalized decoding for pixel, image, and language.
\newblock In \emph{Proceedings of the IEEE/CVF Conference on Computer Vision
  and Pattern Recognition}, pp.\  15116--15127, 2023.

\end{thebibliography}
\bibliographystyle{icml2025}

\newpage

\clearpage
\appendix
\section{Appendix}
\startcontents[sections]
\printcontents[sections]{l}{1}{\setcounter{tocdepth}{2}}

\subsection{Evaluation on 3D Instance Segmentation}
\label{app:3d_instance_seg}
\begin{table*}[h]

\caption{\textbf{Evaluation on 3D Instance Segmentation Benchmarks.} (S) and (M) denotes models trained on sensor and mesh point clouds respectively.}
\label{tab:established_3d}
\begin{subtable}[t]{0.49\linewidth}
\small
\centering
\caption{\textbf{ScanNet200}}
\label{tab:scannet200_ins}
\resizebox{\textwidth}{!}{%
\begin{tabular}{llcc}
\midrule
 & Model  & mAP  & mAP25 \\ \midrule
\multirow{6}{1.2cm}{Closed Vocabulary}
& Mask3D \cite{mask3d} (S) & 15.5  & 24.3  \\
& Mask3D \cite{mask3d} (M) & 27.4   & 42.3 \\
& PQ3D (closed) \cite{pq3d} (M) & 27.0  & 46.3 \\
& QueryFormer \cite{queryformer} (M) & 28.1  & 43.4 \\
& MAFT \cite{maft} (M) & 29.2  & 43.3 \\ 
& ODIN \cite{odin} (S) &  \textbf{31.5}                &  \textbf{53.1} \\  
\midrule 
\multirow{3}{1.2cm}{\revision{Language-Prompted}}
& PQ3D (open) \cite{pq3d} (M) & 20.2  & 32.5 \\
 & \model{}-3D-only (Ours) (S) &   27.9    & 46.1 \\ 
  & \model{} (Ours) (S) &   \textbf{31.0}    & \textbf{52.1} \\  \midrule
\end{tabular}
}
\end{subtable}
\hspace{\fill}
\begin{subtable}[t]{0.50\textwidth}
\small
\centering
\caption{\textbf{Matterport3D}}
\label{tab:matterport_ins}
\resizebox{\textwidth}{!}{%
\begin{tabular}{llcc}
\midrule
Input & Model & mAP & mAP25 \\ \midrule
\multirow{3}{1.2cm}{Closed Vocabulary} & Mask3D \cite{mask3d} (S) & 2.5 & 10.9 \\
  & Mask3D \cite{mask3d} (M) & 11.3 & 23.9 \\
 & ODIN \cite{odin}  (S) &  \textbf{14.5} &  \textbf{36.8} \\
  \midrule
\multirow{2}{1.2cm}{\revision{Language-Prompted}}
 & \model{} (Ours) (S) & 10.7  & 24.5 \\
 & \model{}-2D-3D (Ours) (S) & \textbf{13.4}  & \textbf{29.6} \\
 \midrule
\end{tabular}
}
\end{subtable}

\end{table*}

We test \model{} on 3D segmentation benchmarks of ScanNet200 \cite{scannet200} and Matterport3D \cite{matterport} for instance segmentation tasks. These benchmarks have a fixed vocabulary of objects (200 classes in ScanNet200 and 160 classes in Matterport3D). SOTA models like ODIN \cite{odin} and Mask3D \cite{mask3d} train and evaluate in this fixed vocabulary setup by predicting a distribution over the fixed set of classes and supervising with softmax losses. PQ3D \cite{pq3d} evaluates in a \revision{language-prompted} setup where they supply object names, one object at a time, and gather predictions for all objects in the vocabulary. They compare with a closed-vocabulary version of their model, and find that their \revision{language-prompted} version is about 7\% worse than their closed vocabulary version due to ambiguities in class names confusing CLIP (eg. ``chair" and ``armchair"; ``table" and ``desk" are different categories in ScanNet200). We follow PQ3D and evaluate our model in the \revision{language-prompted} setup. The input to the model is a concatenation of all object classes of the benchmark as a long sentence (eg: ``chair. table. sofa. bed. ...."). While PQ3D cannot predict multiple object classes simultaneously, and hence have to supply one object at a time, our model can simulatenously decode masks for all objects mentioned in the sentence. The results are shown in Table-\ref{tab:established_3d} on the official validation splits of these benchmarks. We observe that \model{} outperforms PQ3D in the \revision{language-prompted} evaluation setup on ScanNet200.

\subsection{Effect of Fine-tuning 2D backbones in \model{}}
We study the effect of fine-tuning the 2D backbones on in-domain and out-of-domain performance. We train two versions of \model{} with swin backbones, one with fine-tuning and the other without fine-tuning. For training, we use SR3D and NR3D, and evaluate on the validation sets of SR3D, NR3D (in-domain) and ScanRefer (out-of-domain). The results of the experiments are shown in Table-\ref{table:finetune}. We find that both models work similarly well, both in-domain and out-of-domain. 

\begin{table}[ht]
    \centering
    \caption{\revision{\textbf{Effect of Fine-tuning 2D backbones of \model{}} for Acc@25 in \Det Setup. SR3D and NR3D are in-domain and ScanRefer is out-of-domain}}
    \label{table:finetune}
    \vspace{6pt}
    \begin{tabular}{p{3cm}ccc}
        \toprule
        \textbf{Model} &  \textbf{SR3D} & \textbf{NR3D} & \textbf{ScanRefer} \\
        \midrule
        \model{} w/ finetune  & 65.6 & 52.7 & 54.4 \\
        \model{} w/o finetune  & 66.7 & 52.0 & 54.5 \\
        \bottomrule
    \end{tabular}
\end{table}

\subsection{Performance with different backbones}
We demonstrate that the performance can scale with the strength of the backbone. Specifically we use a DINOv2~\cite{oquab2024dinov2learningrobustvisual} backbone consisting of 1.1B parameters, scaling over 5x compared to the Swin backbone. To achieve high-performance during training, we freeze the backbone, although we note that it is possible that additional performance could be obtained with efficient fine-tuning techniques such as LoRA~\cite{hu2021lora}. In Table-\ref{table:ablation_backbone}, we find that adding this backbone boosts performance on all 3 language grounding datasets, with substantial margins of $4.5\%$, $1.9\%$, and $3.4\%$ @ 0.25 on SR3D, NR3D, and ScanRefer respectively. 

\begin{table*}[h!]
    \centering
    \caption{\revision{Ablation of visual backbones on 3D language grounding. We evaluate top-1 accuracy on the official validation set without assuming ground-truth proposals (\Det).}}
    \begin{tabular}{l*{12}{c}}
        \midrule
        Method & \multicolumn{3}{c}{SR3D} & \multicolumn{3}{c}{NR3D} & \multicolumn{3}{c}{ScanRefer}\\
        & \begin{tabular}{@{}c@{}}Acc \\ @25\\ (\Det)\end{tabular} & \begin{tabular}{@{}c@{}}Acc \\ @50\\ (\Det)\end{tabular} & \begin{tabular}{@{}c@{}}Acc \\ @75\\ (\Det)\end{tabular} & \begin{tabular}{@{}c@{}}Acc \\ @25\\ (\Det)\end{tabular} & \begin{tabular}{@{}c@{}}Acc \\ @50\\ (\Det)\end{tabular} & \begin{tabular}{@{}c@{}}Acc \\ @75\\ (\Det)\end{tabular} &  \begin{tabular}{@{}c@{}}Acc \\ @25\\ (\Det)\end{tabular} & \begin{tabular}{@{}c@{}}Acc \\ @50\\ (\Det)\end{tabular} & \begin{tabular}{@{}c@{}}Acc \\ @75\\ (\Det)\end{tabular} \\
        \midrule
        \model\;(Swin) & 67.1 & 58.7 & 46.4  & 50.6 & 41.9 & 32.2 & 57.3 & 49.8 & 40.2 \\
         \model\;(DINOv2) & 71.6 & 63.8 & 49.4 & 52.5 & 43.3 & 34.2 & 60.7 & 53.2 & 42.6 \\
        \midrule 
    \end{tabular}
    \label{table:ablation_backbone}
\end{table*}

\subsection{Additional Metrics on ScanQA Dataset}
We report additional standard metrics used by ScanQA benchmark in Table-\ref{table:vqa_detail}.

\subsection{Visualizations of \model{} on  Referential Grounding Datasets}
\begin{figure*}[ht]
\centering
    \includegraphics[width=\textwidth]{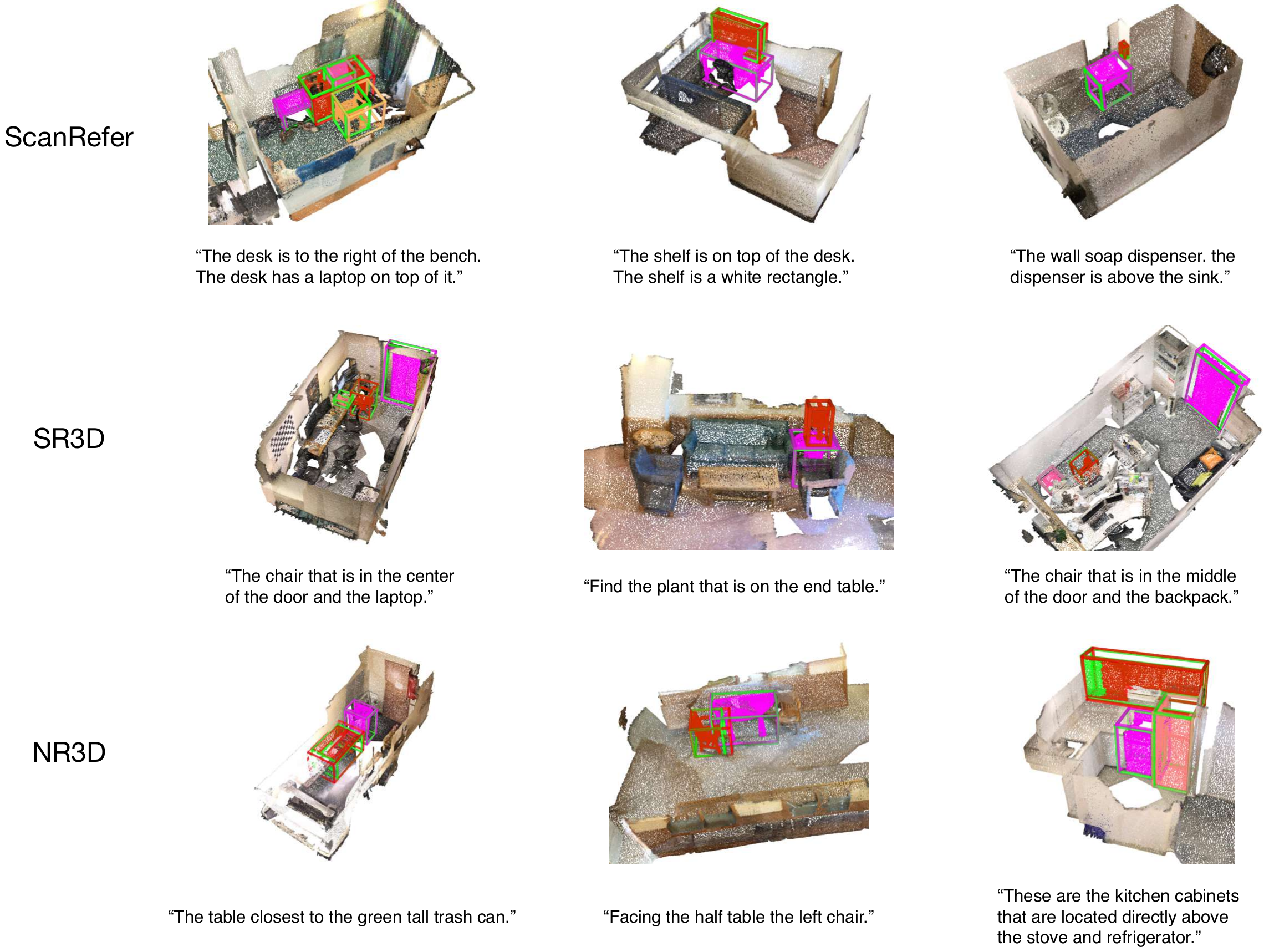}
    \caption{\textbf{Visualizations of \model{} on 3D Referential Grounding Datasets of ScanRefer, SR3D, and NR3D} The red masks indicate \model{}'s prediction for the target object, and pink masks indicate its predictions for the anchor object. Green masks and boxes indicate ground-truth target and anchor objects.
    }
    \label{fig:vis}
\end{figure*}

\begin{figure*}[ht]
\centering
    \includegraphics[width=\textwidth]{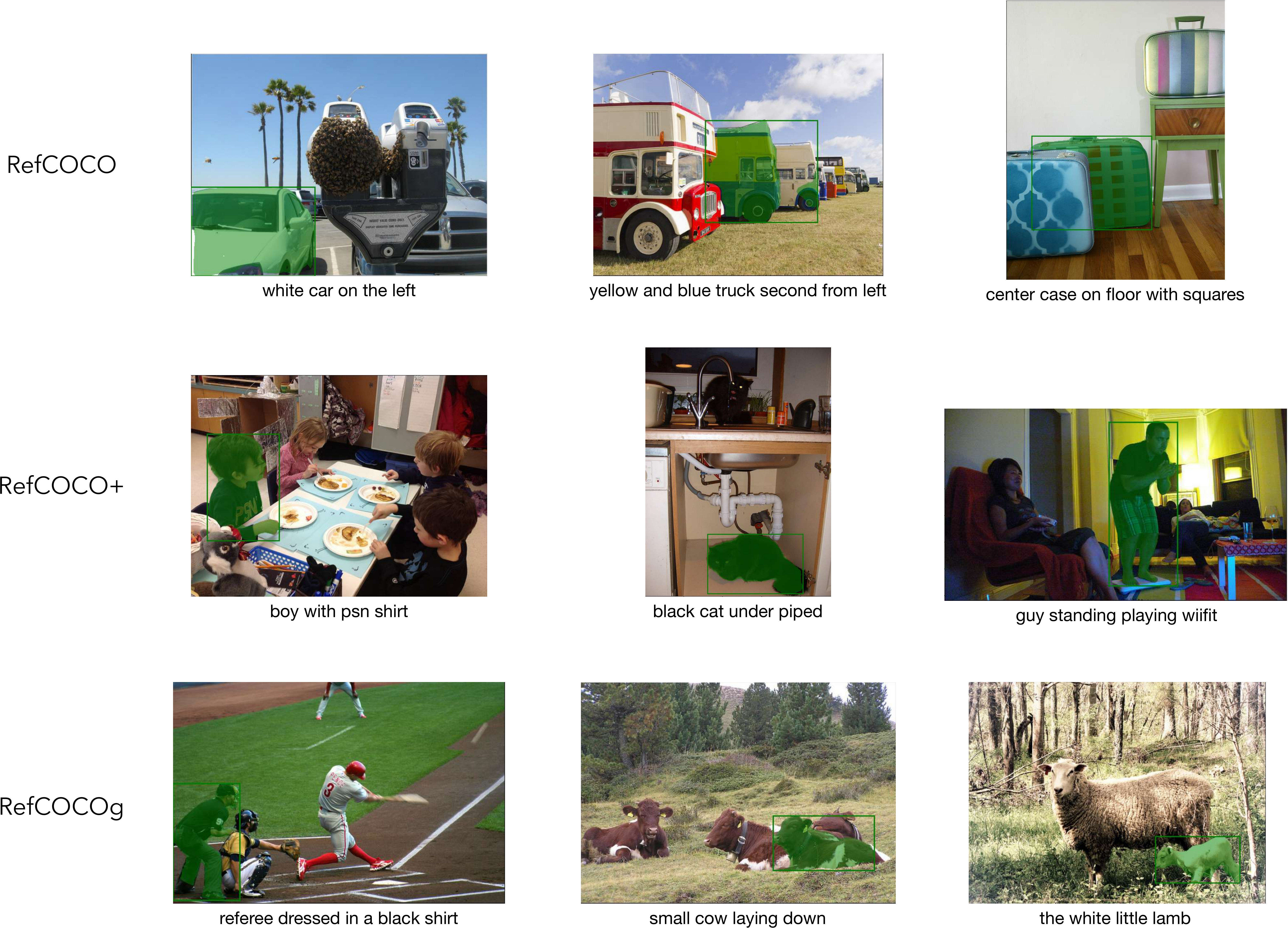}
    \caption{\textbf{Visualizations of \model{} on 2D Referential Grounding Datasets of RefCOCO, RefCOCO+, and RefCOCOg}: The green masks indicate predictions of \model{}.}
    \label{fig:vis_refcoco}
\end{figure*}

We show the visualization of \model{} on 3D referential grounding in Figure-\ref{fig:vis} and on 2D referential grounding in Figure-\ref{fig:vis_refcoco}.

\begin{figure*}[ht!]
\centering
    \includegraphics[width=\textwidth]{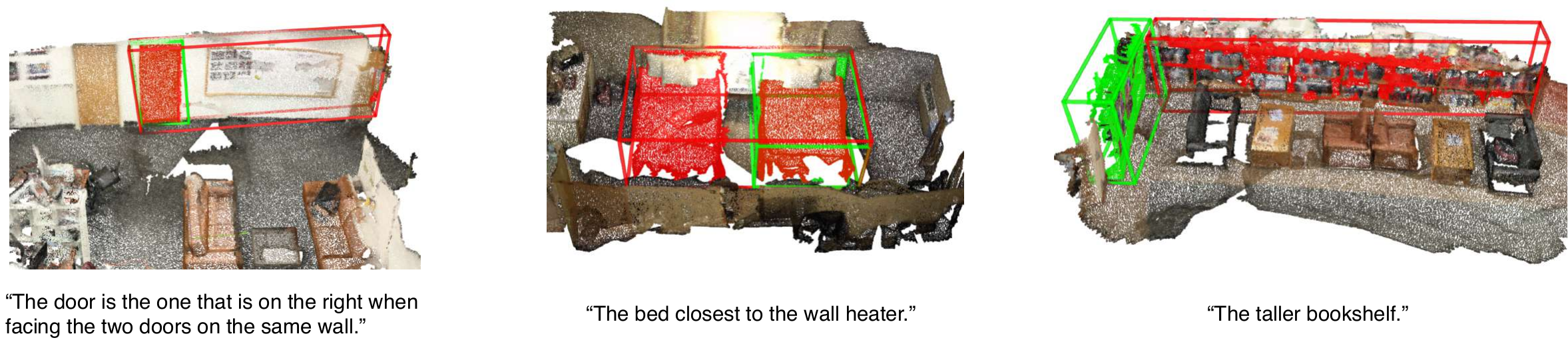}
    \caption{\textbf{Systematic failure modes of \model{}}: Green boxes and masks are ground-truth, red masks and boxes are \model{}'s predictions. 
    }
    \label{fig:failure}
\end{figure*}

We identify three systematic failure modes in our model, illustrated in Figure-\ref{fig:failure}. 
\begin{itemize}
    \item \textbf{Inclusion of distant outlier points in the predicted masks}: In the first image of Figure-\ref{fig:failure}, while \model{} accurately predicts the object, it also mistakenly includes some distant points in the mask. This leads to a larger bounding box during the mask-to-bounding box conversion in post-processing, negatively affecting accuracy metrics. Although our proposed box loss mitigates this issue, it doesn't fully resolve it. 
    \item \textbf{Multiple instances of the same object being segmented together}: As shown in the middle image of Figure-\ref{fig:failure}, \model{} predicts both beds as a single output. Incorporating attention to language and queries helps reduce such errors, though they still persist. Our box loss also aids in addressing this issue.
    \item \textbf{Failures in language understanding} as seen in the third image of Figure-\ref{fig:failure}.     
\end{itemize}

The first two failure modes are specific to mask-decoding architectures, and similar issues have been noted by Mask3D \cite{mask3d} in their 3D instance segmentation tasks. Box-decoding architectures, on the other hand, generally avoid these problems. Nevertheless, we find that mask-decoding architectures offer significant advantages in other aspects, such as more accurate and fine-grained segmentation, making them valuable despite these challenges.

\begin{table*}[ht]
    \centering
    \caption{\textbf{Extra Metrics on ScanQA validation set}}
    \vspace{2pt}
    \label{table:vqa_detail}
    \revision{\begin{tabular}{llccccc}
        \toprule
        & \textbf{Method} & \textbf{EM} & \textbf{BLEU-1} & \textbf{ROUGE} & \textbf{METEOR} & \textbf{CIDEr}\\
        \midrule
        \multirow{4}{*}{Mesh PC}
        & 3D-LLM~\cite{3dllm} & 20.5 & \textbf{39.3} & 35.7 & 14.5 & 69.4 \\
        & PQ3D~\cite{pq3d} & 21.0 & - & - & - & - \\
        & 3D-VisTA~\cite{3dvista} & 22.5 & 32.0 & 35.5 & 13.8 & 69.1 \\
        & NaviLLM~\cite{navillm} & \textbf{23.0} & - & \textbf{38.4} & \textbf{15.4} & \textbf{75.9} \\
        \midrule
        \multirow{2}{*}{Sensor PC}
        & 3D-VisTA~\cite{3dvista} & 21.6 & 30.1 & 34.1 & 13.2 & 65.3 \\
        & \model{} (Ours) & \textbf{25.7} & \textbf{36.1} & \textbf{40.0} & \textbf{15.2} & \textbf{78.5} \\
        \bottomrule
    \end{tabular}}
\end{table*}

\begin{figure*}[ht!]
\centering
    \includegraphics[width=\textwidth]{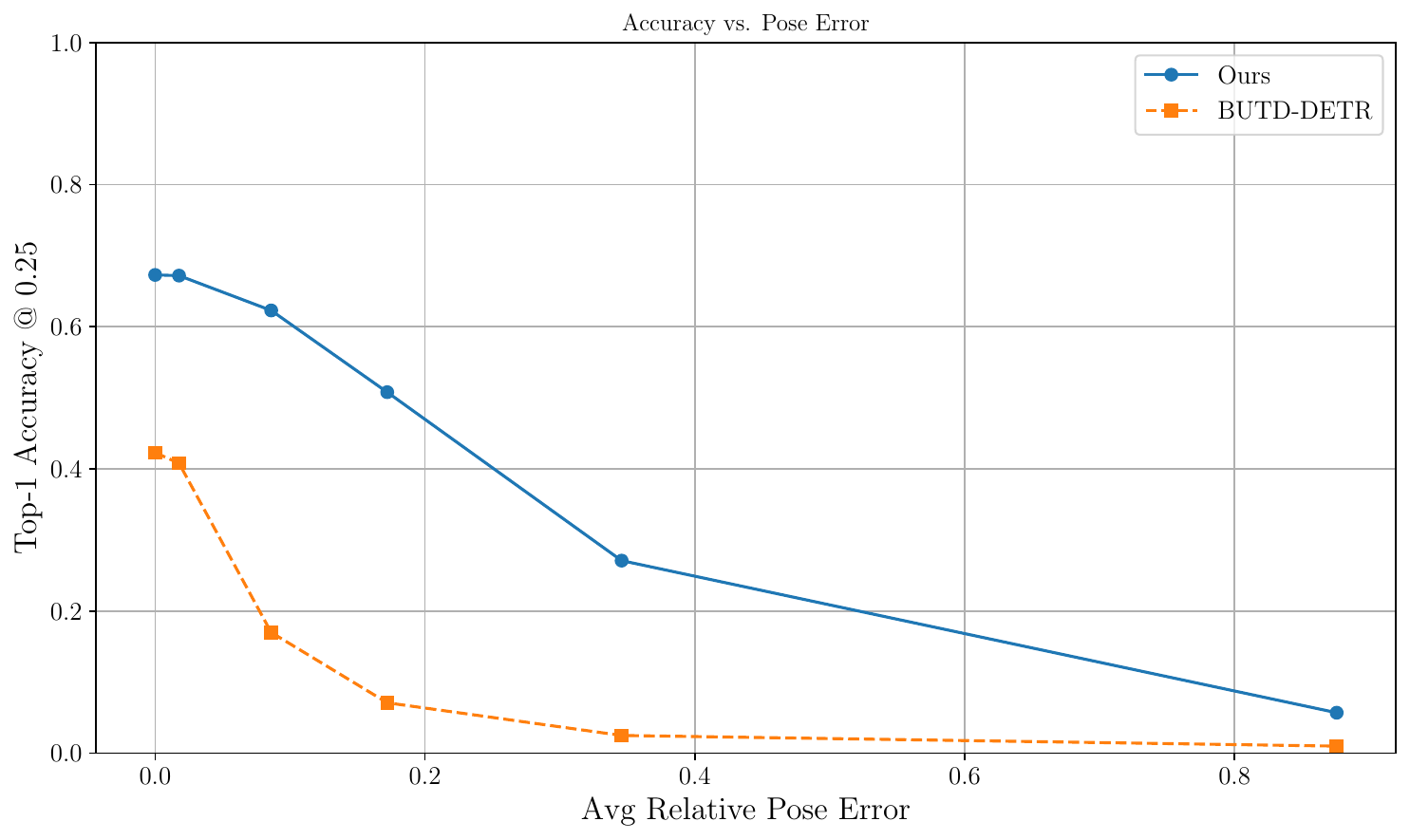}
    \includegraphics[width=\textwidth]{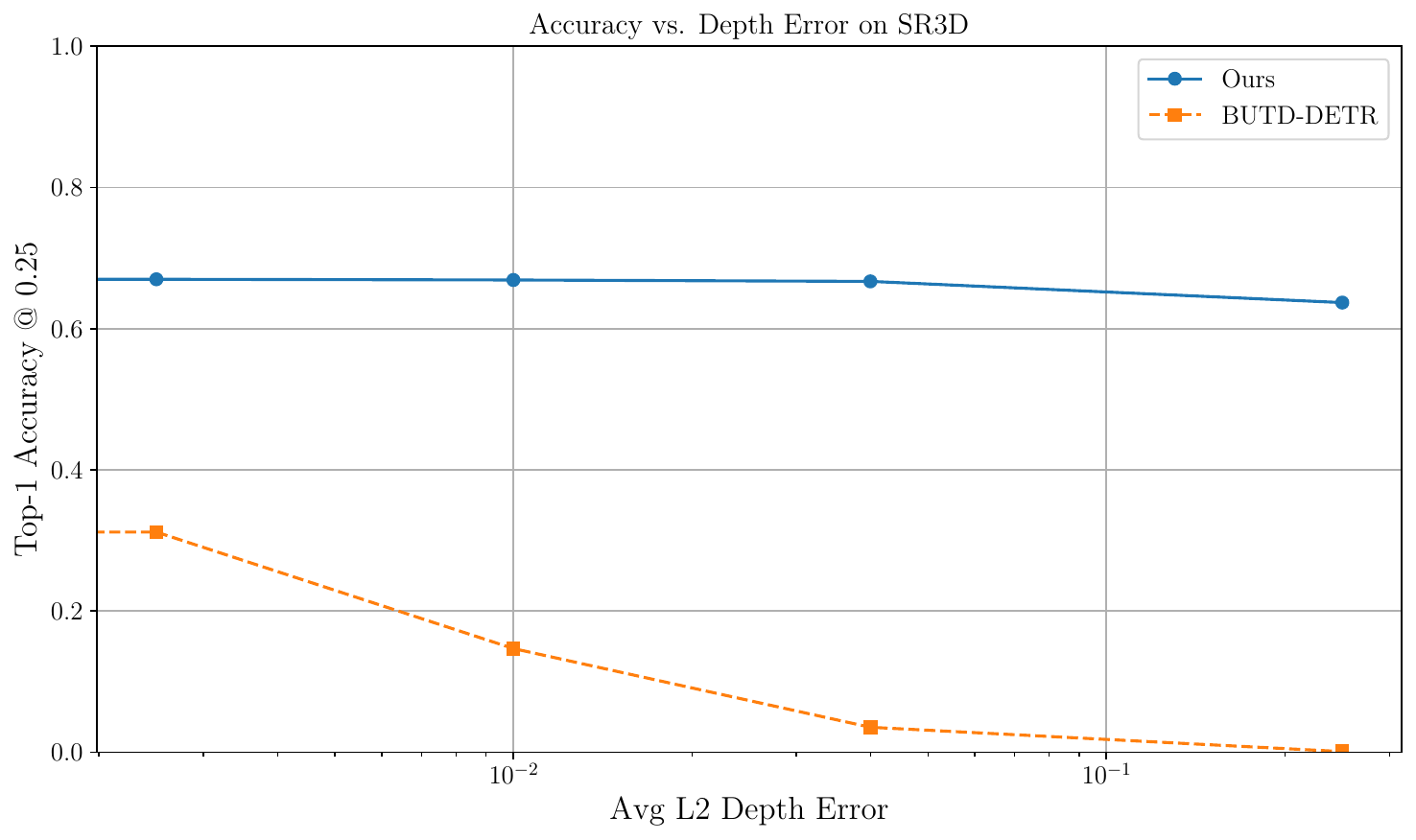}
    \caption{\revision{We analyze the performance of \model{} and BUTD-DETR on SR3D as the pose and depth error increases. We add gaussian noise to the pose and raw depth which affects the unprojected point cloud that both models observe.}}
    \label{fig:rodin_noise}
\end{figure*}

\subsection{Performance analysis with pose and depth noise}
To analyze the performance of \model{} under sensor noise we conduct two experiments to model error in both pose and depth. For the pose error experiment, we add gaussian noise to the translation and rotation components of every camera pose in a scene. Similarly, for the depth error experiment, we add gaussian noise uniformly to the depth map. When each depth map is unprojected, the resulting point cloud becomes misaligned and performance decreases. We use relative pose error as defined in~\cite{6385773}. 

We compare the robustness of \model{} to prior state-of-the-art single-stage method of BUTD-DETR \cite{butd}. We chose a single-stage method as our baseline, since multi-stage methods like PQ3D \cite{pq3d} and 3D-VisTA \cite{3dvista} rely on several external models, and use pre-processed intermediate outputs from them for their inference. This makes it harder to fairly run comparisons directly on the point cloud input.  
As shown in Figure-\ref{fig:rodin_noise}, \model{} is highly robust to both types of noise. 
At a mean error of $0.2$, \model{} impressively maintains a Top1@0.25 IoU accuracy of $66.7\%$.

In the pose error case, the model must understand the misaligned point cloud and cannot simply ignore the spurious points. However, \model{} still shows impressive robustness with substantially less degradation compared to BUDT-DETR.

We believe a great portion of robustness comes from reliance on 2D pre-trained features and 2D layers in the network. Despite the noise in the depth and pose estimation, they still operate over the clean RGB images. Additionally, our 3D layers use local and relative attentions, which additionally contribute to the robustness.

\subsection{Miscellaneous Details}
\paragraph{Frame Sampling}
To improve training efficiency, we opt to train on only a subset of available frames in each scene. This is critical, not only for reducing the average cost per step, but also in ensuring that the computations and memory is fixed per-step, allowing us to maximize the batch size and prevent waiting between GPUs in DDP. We initially tested a random selection strategy where each caption was paired with $N=15$ images from a given scene, with each scene originally containing around $90$ frames. However, this change means that some sets of frames may no longer align with a given caption (i.e., the referenced object may not be visible in the selected frames). By simply ignoring instances where this mismatch occurs, this strategy performed remarkably well overall.

However, to improve performance further, we sought to specifically include frames that were relevant to the caption (e.g., given "the red chair", we want to make sure all chairs in the scene are included), without biasing the model and causing a train/test distribution gap. To do this, we computed the CLIP embeddings of all scene frames, as well as the text embeddings of all captions. For a given caption, we select $5$ relevant frames using the cosine similarity of these embeddings, ensuring that referenced entities are part of the selected frames. Moreover, to ensure diversity of frames and prevent the aforementioned bias, we select the remaining $10$ frames using Furthest-Point-Sampling (FPS) on the CLIP image embedding space. We found fewer than 15 frames to slightly reduce performance, but that increasing beyond this point had no benefit.

\paragraph{Language Encoder}
As the original CLIP model only supported a maximum of 77 tokens, we opted to use Jina-CLIP which uses RoPE embeddings and is trained on 512 tokens.

\paragraph{Matching Cost} Tuning the Hungarian matching cost weights turned out to be critical for 3D referential grounding. In prior works \cite{odin, m2f}, mask cost weights are typically higher than class cost weights. We find that it is important to have high class cost weights in referential grounding. This difference in costs is only required in Hungarian matching, we do not observe much benefits in changing the loss weights from what was used in the prior works.

\paragraph{Mask Decoder design} Unlike prior works~\cite{odin, m2f}, where object queries attend to visual features across multiple resolutions, we attend only to the highest-resolution feature scale. We explored multi-scale variants of our model but struggled to achieve good results, primarily due to difficulties in properly updating multi-scale visual feature maps with object queries and language tokens. Single-scale mask decoding approach is suboptimal for backbones like Swin, where attending to high-resolution feature maps is computationally expensive. However, with ViT backbones, where all feature maps share the same resolution, this limitation is less pronounced.

\paragraph{Box Loss} As described in the main text, we employ the box loss to enhance the sharpness of the predicted masks. The box loss penalizes outlier points predicted by the mask head, which might not receive significant penalties through the mask loss alone. We incorporate the box loss into both Hungarian matching and the final loss used for backpropagation. Initially, we hypothesized that applying the box loss solely in Hungarian matching would suffice, as the bounding box is derived only from the extreme points of the predicted mask, resulting in sparse gradients during backpropagation. However, empirical results demonstrate that explicitly including the box loss in the final loss is beneficial. Notably, we use bounding box loss exclusively for 3D datasets and not for (lifted) 2D datasets. This distinction arises because lifted 3D point clouds can be noisy, potentially leading to inaccurate 3D bounding boxes.

\subsection{Discussion: Decoding Masks vs. Boxes}
Decoding 3D bounding boxes has its advantages. For instance, datasets like Arkit3DScenes~\cite{arkitscenes} and Aria~\cite{straub2024efm3dbenchmarkmeasuringprogress} only provide supervision for 3D boxes, making box decoding more favorable in such scenarios. However, recent methods such as Box2Mask~\cite{box2mask} demonstrate that segmentation predictions can be effectively supervised using bounding box annotations alone, suggesting that the lack of mask labels may not pose a significant limitation. Despite this, we observe failure cases in mask predictions, such as outlier points being segmented in 3D or multiple instances of the same object being predicted as the answer (see Figure~\ref{fig:failure} in Appendix). These errors result in oversized or incorrect bounding boxes when masks are converted to boxes in a post-processing step, indicating the need for further research to address these issues in mask-decoding heads.

On the other hand, decoding masks offers significant benefits. As our experiments show, predicting masks leads to better performance compared to decoding boxes. Moreover, models like ODIN and \model{} aim to unify 2D and 3D perception tasks, and mask prediction provides a consistent interface across these modalities. Masks represent per-point segmentation of pixels or points, whether in 2D or 3D, whereas box decoding requires separate prediction heads for 2D (4D outputs) and 3D (6D outputs), complicating unification efforts. Additionally, box-decoding heads are sensitive to normalization requirements for 3D scenes, such as mean-centering or scaling within a specific range (e.g., 0 to 1). Indoor and outdoor datasets, which vary significantly in 3D extents, often require separate decoder heads, further complicating training. In contrast, mask-decoding heads rely on cosine similarity between pixel/point features and object queries, making them more robust to variations in scene normalization and context. For example, \model{} maintains strong performance even when input scenes are translated by 1000 meters, despite not being explicitly trained with translation augmentations. Furthermore, annotating 3D masks is easier for humans than annotating 3D boxes, especially with recent advances in interactive segmentation methods like SAM~\cite{sam}, which simplify the process of creating accurate mask annotations. This robustness, simplicity, and annotation efficiency make mask-decoding heads a preferable choice for unifying 2D-3D perception tasks.

\begin{table*}[h!]
    \centering
    \caption{\revision{Mask mAP evaluation on 3D language grounding. We evaluate top-1 accuracy on the official validation set without assuming ground-truth proposals (\Det).}}
    \begin{tabular}{l*{12}{c}}
        \midrule
        Method & \multicolumn{3}{c}{SR3D} & \multicolumn{3}{c}{NR3D} & \multicolumn{3}{c}{ScanRefer}\\
        & \begin{tabular}{@{}c@{}}Acc \\ @25\\ (\Det)\end{tabular} & \begin{tabular}{@{}c@{}}Acc \\ @50\\ (\Det)\end{tabular} & \begin{tabular}{@{}c@{}}Acc \\ @75\\ (\Det)\end{tabular} & \begin{tabular}{@{}c@{}}Acc \\ @25\\ (\Det)\end{tabular} & \begin{tabular}{@{}c@{}}Acc \\ @50\\ (\Det)\end{tabular} & \begin{tabular}{@{}c@{}}Acc \\ @75\\ (\Det)\end{tabular} &  \begin{tabular}{@{}c@{}}Acc \\ @25\\ (\Det)\end{tabular} & \begin{tabular}{@{}c@{}}Acc \\ @50\\ (\Det)\end{tabular} & \begin{tabular}{@{}c@{}}Acc \\ @75\\ (\Det)\end{tabular} \\
        \midrule
         \model\; & 75.4 & 69.2 & 54.4 & 59.9 & 52.3 & 39.4 & 66.8 & 60.8 & 48.0 \\
        \midrule 
    \end{tabular}
    \label{table:mask_iou_eval}
\end{table*}

\subsection{Evaluation with Mask based Intersection Over Union} While standard evaluations on existing 3D language grounding benchmarks rely on bounding box Intersection Over Union (IoU) for computing accuracy, we additionally report results using mask-based IoU for accuracy computation. The corresponding results are presented in \cref{table:mask_iou_eval}.

\subsection{Additional Related Work}
\label{appendix:related_work}
\paragraph{Language Understanding Benchmarks}
Vision Language Grounding is the task of localizing  the objects mentioned in a language utterance in a given 2D or 3D scene. In the 2D domain, this task is primarily benchmarked on RefCOCO/+/g datasets \cite{refcoco} with humans annotating language instructions on top of COCO images. In the 3D community, this task is primarily studied in the popular benchmarks of SR3D \cite{referit3d} containing programmatically generated sentences, and NR3D \cite{referit3d} and ScanRefer \cite{scanrefer}, containing human-annotated sentences, and 3D scenes from the ScanNet \cite{scannet} dataset. The original benchmarks of SR3D and NR3D provide access to ground-truth bounding boxes of all objects in the scenes as input, and the task is to select the correct bounding box that corresponds to the language sentence. Most methods operate under this assumption, except for BUTD-DETR \cite{butd}, which proposed directly predicting  3D bounding boxes instead of selecting from the available proposals. We follow BUTD-DETR and report results without assuming access to ground-truth boxes. The ScanRefer benchmark is similar to NR3D but does not provide ground-truth boxes as input. Recently, ScanQA \cite{scanqa} and SQA3D \cite{sqa3d} introduced 3D Question Answering Benchmarks. ScanQA focuses on spatial relations. SQA3D \cite{sqa3d} provides pairs of situation descriptions and questions regarding embodied scene understanding, navigation, common sense and multi-hop reasoning.

\paragraph{3D Question Answering and Captioning} For 3D question answering and captioning, approaches like PQ3D \cite{pq3d} and 3D-Vista \cite{3dvista} use small text generation heads on top of their language-contextualized features or queries to decode answers. Other approaches like 3D-LLM \cite{3dllm} and NaviLLM \cite{navillm} condense the visual scene features into a set of latent vectors and pass it to large pre-trained LLMs like BLIP2-flant5~\cite{blip2} or Vicuna-7B-v0 \cite{vicuna}. However, unlike 3D-Vista and PQ3D, they either get significantly poor performance on 3D referential grounding tasks (3D-LLM) or skip evaluating in that setup (NaviLLM). In this work, we follow PQ3D and 3DVista's approach and use a small text generation head, mainly for its simplicity.

\paragraph{Sensor vs Mesh Point Clouds in 3D benchmarks: } All the 3D benchmarks use point clouds derived from the 3D meshes provided by ScanNet~\cite{scannet}. These meshes were constructed using several steps of post-processing over the raw sensor RGB-D data (which takes minutes-to-hours). In this work, we propose the first 3D language grounding model that operates directly over sensor RGB-D point clouds. For fair comparison, we benchmark other prior works with sensor point clouds as inputs, and show the benefits of using 2D pre-trained features for 3D language understanding tasks. These post-processing steps include mesh reconstruction and camera pose estimation, as well as several manual post-processing steps. These processes create fine-grained misalignments between the reconstructed mesh and the sensor RGB-D stream, resulting in drop in performance for methods operating over sensor RGB-D streams instead of the mesh point clouds, as also shown by prior works \cite{deepview,virtualview,odin}. This discourages the use of sensor RGB-D streams and thus the 2D features pre-trained on internet scale data.  Using sensor point clouds directly is an emerging idea in the community, further bolstered by the recent introduction of datasets like EmbodiedScan \cite{embscan} which also use sensor data directly instead of using meshes.

\subsection{Detailed ReferIt3D Results} We provide detailed results for the ReferIt3D benchmark in~\cref{tab:detailed_referit}.

\begin{table*}[h!]
   \centering
  \small
  \setlength{\tabcolsep}{2pt}
  \renewcommand{\arraystretch}{0.85}
  \caption{Detailed results ReferIt3D on \textbf{Nr3D} and \textbf{Sr3D}.\vspace{8pt}}
  \makebox[\textwidth][c]{%
    \begin{tabular}{l*{10}{c}}
      \toprule
      \textbf{Method}
        & \multicolumn{5}{c}{\textbf{Nr3D}}
        & \multicolumn{5}{c}{\textbf{Sr3D}} \\
      \cmidrule(lr){2-6} \cmidrule(lr){7-11}
        & \textbf{Overall}
        & \textbf{Easy}
        & \textbf{Hard}
        & \textbf{View-Dep}
        & \textbf{View-Indep}
        & \textbf{Overall}
        & \textbf{Easy}
        & \textbf{Hard}
        & \textbf{View-Dep}
        & \textbf{View-Indep} \\
      \midrule
      ViL3DRel~\cite{vil3d}
        & 64.4  & 70.2 & 57.4  & 62.0  & 64.5
        & 72.8  & 74.9 & 67.9 & 63.8 & 73.2 \\
      CoT3DRef~\cite{abdelrahman2024cot3drefchainofthoughtsdataefficient3d}
        & 64.4  & 70.0 & 59.2  & 61.9  & 65.7
        & 73.2  & 75.2 & 67.9 & 67.6 & 73.5 \\
      MiKASA~\cite{chang2024mikasa}
        & 64.4  & 69.7 & \textbf{59.4} & \textbf{65.4} & 64.0
        & 75.2  & 78.6 & 67.3 & \textbf{70.4} & 75.4 \\
      3D-VisTA~\cite{3dvista}
        & 64.2  & 72.1 & 56.7  & 61.5  & 65.1
        & 72.8  & 74.9 & 67.9 & 63.8 & 73.2 \\
      \midrule
      \model{}
        & \textbf{65.2} & \textbf{73.3} & 57.0  & 55.1  & \textbf{69.9}
        & \textbf{81.7} & \textbf{84.4} & \textbf{75.2} & 66.2  & \textbf{82.4} \\
      \bottomrule
    \end{tabular}%
  }
  \label{tab:detailed_referit}
\end{table*}

\end{document}